\documentclass[sigconf]{acmart}

\usepackage{caption}
\usepackage{subcaption}

\AtBeginDocument{%
  }

\setcopyright{none}
\renewcommand\footnotetextcopyrightpermission[1]{}
\acmConference[]{}{}{}

\begin{document}

\title{Mutation Models:\\ Learning to Generate Levels by Imitating Evolution}
\author{Ahmed Khalifa}
\email{ahmed@akhalifa.com}
\affiliation{%
  \institution{Institute of Digital Games\\University of Malta}
  \city{Msida}
  \country{Malta}
}

\author{Michael Cerny Green}
\email{mike.green@nyu.edu}
\affiliation{%
  \institution{Game Innovation Lab\\New York University}
  \city{Brooklyn}
  \state{NY}
  \country{USA}
}

\author{Julian Togelius}
\email{julian@togelius.com}
\affiliation{%
  \institution{Game Innovation Lab\\New York University}
  \city{Brooklyn}
  \state{NY}
  \country{USA}
}

\renewcommand{\shortauthors}{Khalifa et al.}

\begin{abstract}
Search-based procedural content generation (PCG) is a well-known method for level generation in games. Its key advantage is that it is generic and able to satisfy functional constraints. However, due to the heavy computational costs to run these algorithms online, search-based PCG is rarely utilized for real-time generation. In this paper, we introduce mutation models, a new type of iterative level generator based on machine learning. We train a model to imitate the evolutionary process and use the trained model to generate levels. This trained model is able to modify noisy levels sequentially to create better levels without the need for a fitness function during inference. We evaluate our trained models on a 2D maze generation task. We compare several different versions of the method: training the models either at the end of evolution (normal evolution) or every 100 generations (assisted evolution) and using the model as a mutation function during evolution. Using the assisted evolution process, the final trained models are able to generate mazes with a success rate of $99\%$ and high diversity of $86\%$. The trained model is many times faster than the evolutionary process it was trained on. This work opens the door to a new way of learning level generators guided by an evolutionary process, meaning automatic creation of generators with specifiable constraints and objectives that are fast enough for runtime deployment in games.
\end{abstract}



\maketitle

\pagestyle{plain}

\section{Introduction}

Very coarsely, we can construct content generators in two different ways. We can either create a generator that constructs an artifact in a finite (often fixed) number of steps without testing during the construction process. The other way is to perform a search or optimization process where either a whole artifact or part of it is tested repeatedly during generation to guide it forward~\cite{togelius2011search}. In general, constructive generators are much faster and therefore better suited to real-time generation than search-based generators. On the other hand, the lack of quality checking during the construction process means that the expressive spaces of the generator may need to be restricted in order to guarantee that the content is playable (not functionally broken). Creating a good constructive generator is hard, and often requires expert human effort for each use case. Generators based on search and optimization, on the other hand, can guarantee playability but are much slower; depending on the cost of content evaluation, they can be unworkably slow.

Could these advantages be combined? Could we create constructive generators that are fast yet have functionality guarantees and a wide expressive range? Could these be created automatically? It stands to reason that one could use machine learning to somehow learn generators. While the computational costs of learning a generator may be large, using the learned model as a constructive generator is computationally cheap; in other words, computation is front-loaded. Various self-supervised approaches have been advanced, building on GANs~\cite{volz2018evolving,torrado2020bootstrapping} or autoencoders~\cite{jain2016autoencoders,sarkar2021dungeon}, that learn from existing content to generate new content. Another recent approach, the Path of Destruction, turns existing content into sequences of repair actions that can be modeled~\cite{siper2022path}. Obviously, self-supervised approaches require a decently-sized library of content, such as levels, to learn from~\cite{summerville2018procedural,liu2021deep}.

\begin{figure*}
    \centering
    \includegraphics[width=0.7\linewidth]{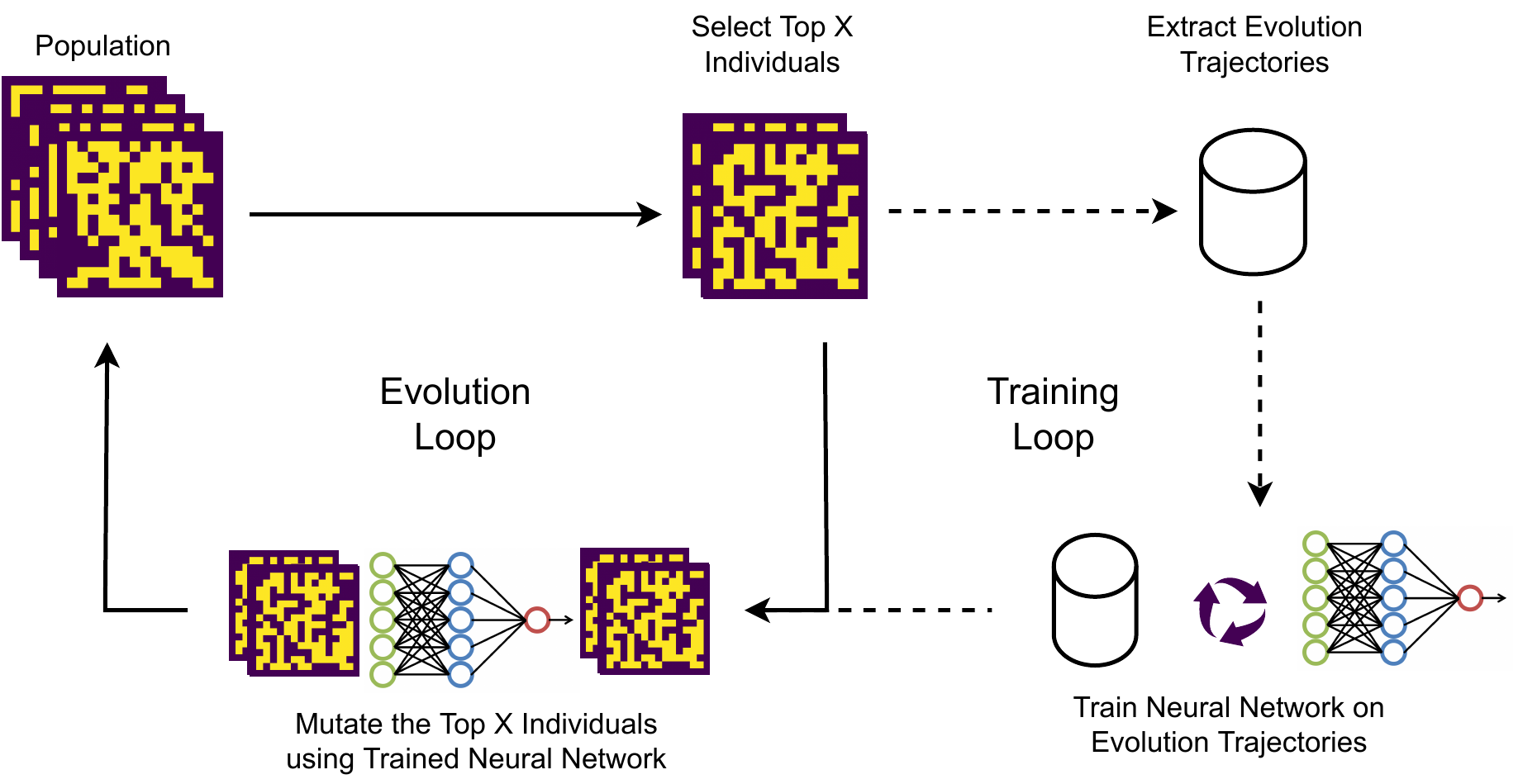}
    \caption{System diagram of the mutation models framework. The system consists of two main parts: evolution loop and training loop. The evolution loop is the usual evolutionary algorithm which evolves chromosomes using a fitness function. The training loop converts the evolution history into a dataset then trains a machine learning model on it. This trained machine learning model can be used to assist evolution by acting as a mutator or only trained at the end of evolution. If the model is being used as a mutator, the training loop happens every $I$ generations to improve the accuracy of the machine learning model. Later when the evolution is done, the trained model can be used as a level generator as it learned to imitate the successful evolution trajectories.}
    \label{fig:system}
\end{figure*}

Alternatively one could use evolution~\cite{kerssemakers2012procedural,khalifa2020multi,earle2022illuminating} or reinforcement learning~\cite{khalifa2020pcgrl,shu2021experience,mahmoudi2021arachnophobia} (PCGRL) to learn generators. This does not require existing content, but it does necessitate the existence of a good reward or fitness function. Specifying a good fitness/reward function for generating content generators can be much harder than specifying a good fitness/reward function for ``simply''  generating content. The intuition behind this is that we more or less know what good content looks like, but not necessarily what a good content \emph{generator} looks like. To take a concrete issue, a good fitness/reward function for a generator will probably need to reward appropriately diverse output from the generator. 

In this paper, we propose another way of generating competent content generators. Much like PCGRL, we are not reliant on existing content examples. However, our method also does not need a way to evaluate content generator quality; we only need a way to evaluate good content. The goal is to make it possible to create fast content generators that are as reliable, controllable, and easy to specify as search-based generators. In a nutshell, our approach entails building a search-based generator, and training a neural network to take the actions the evolutionary algorithm would take. The sequence of changes that are made by evolution can be seen as training data for a supervised learning process, and the resulting model will take those actions that the evolutionary process would. This can be seen as imitation learning on evolutionary trajectories; the approach has similarities with offline reinforcement and surrogate models in evolution.

\section{Background}
In the following subsections, we review related background work in regards to procedural content generation (PCG), focusing on search-based and machine learning based methods. We also review the concept of surrogate modeling for evolutionary systems.

\subsection{Search-based PCG}
Search-based PCG defines a family of PCG techniques powered by search methods to generate content~\cite{togelius2011search}. Evolutionary algorithms are often used, as they can be applied to many problems like level generation in video games. Search-based PCG has been applied to many frameworks and games, such as the General Video Game AI framework~\cite{perez2019general}, PuzzleScript~\cite{khalifa2015automatic}, Cut the Rope (ZeptoLab, 2010)~\cite{shaker2013evolving}, and Mazes~\cite{ashlock2011search}. Search-based techniques have been used to generate levels~\cite{ashlock2010elements,khalifa2015automatic,bhaumik2020tree}, game rules~\cite{browne2010evolutionary,khalifa2017general,cook2013mechanic} and level generators~\cite{kerssemakers2012procedural,khalifa2020multi,earle2022illuminating}.

\subsection{PCGML}
Procedural content generation via machine learning (PCGML) is a process of generating content using ML algorithms based on input example. These methods are not often used outside of the research community because of their reliance on large datasets, long training times, and little control of the generated output. There are exceptions: for example, Caves of Qud~\cite{griblat2016caves} uses PCGML to generate books and other aesthetic elements. Research applications with PCGML have used many different techniques including Markov Chains~\cite{snodgrass2016learning}, N-Grams~\cite{dahlskog2014linear}, GANs~\cite{volz2018evolving,torrado2020bootstrapping}, Autoencoders~\cite{jain2016autoencoders,sarkar2021dungeon}, and LSTMs~\cite{summerville2016learning}.

Most related to this work is a project by Siper et al. in which a network is trained to repair levels~\cite{siper2022path}, modeling a so-called \textit{Path of Destruction}. This can be done by taking a series of ``goal levels'' and proceeding to randomly destroy them. A network is then trained on the repair trajectories to learn how to convert these destroyed levels back to the goal levels. The difference between Path of Destruction and most of the recent PCGML work is the network generates the content iteratively (similar to Wave Function Collapse~\cite{karth2017wavefunctioncollapse} and Markov Random Fields~\cite{snodgrass2016learning}) while for example GAN-based or autoencoder-based generators generate content in one shot (one pass). The approach proposed is similar to Path of Destruction in that we are training a networks to iteratively improve randomized levels, but instead of training it on reversed paths of destruction we train it by imitating the successful mutation operation of evolution.

\subsection{PCGRL}
Procedural content generation via reinforcement learning (PCGRL)~\cite{khalifa2020pcgrl} is a process of generating content using RL algorithms. PCGRL transforms the generation process to a game playing process where an agent can take actions at several states and get a reward based on that. The work proposes three different methods to achieve that: narrow, turtle, and wide representations. Narrow representation transforms the generation into a process of asking the agent about each tile in the level if it needs to change or not and if yes what is the value of change. On the other hand, turtle representation has more control over the location by allowing the agent to either modify the current tile (like narrow) or move to a neighboring tile. Finally, wide representation provides the agent with maximum control by allowing the agent to select the location freely and the modification value for it. In this work, we will be using the narrow representation as it provides a small action space for the agent and has comparable results to the other two~\cite{khalifa2020pcgrl}.

PCGRL techniques have been used in level/experience generation~\cite{chen2018q,nam2019generation,werneck2020generating,earle2021learning,zakaria2022procedural,mahmoudi2021arachnophobia,shu2021experience}. It is difficult to transform content generation into a reinforcement environment, thus there is a few PCGRL examples in either academia or industry compared to other methods. Similar to Path of Destruction, PCGRL generators are \emph{iterative}, meaning that they produce content using multiple modification steps, action by action. This grants certain advantages to PCGRL: for example, it is often easier to build mixed-initiative systems around them~\cite{guzdial2018co,guzdial2019friend,delarosa2021mixed}. 

\subsection{Surrogate Modeling}
Surrogate models are models of computationally expensive processes, which can then be used in lieu of the process itself~\cite{ong2003evolutionary}. Surrogate models are meant to be easier to evaluate than the process it models, which translates to time and computational improvements. Surrogates are typically constructed using a data-driven, bottom-up approach, typically by training a network on a distribution of intelligently selected data points.

Within evolutionary computation, it is common to use machine learning to build surrogate models of the fitness function~\cite{jin2011surrogate}. A fitness surrogate model takes a genome as input and returns a fitness value, just as the fitness function would; the advantage is that the trained model is much faster than the actual fitness function. In evolutionary algorithm applications where the fitness function commonly dominates the computation time, surrogate modeling is very useful. These surrogate models can be trained during an individual evolutionary run or over several such runs. While it would be possible in principle to learn surrogate models of all parts of an evolutionary algorithm, they are predominantly applied to the fitness function. The only attempt we know to create surrogate-based mutation is built on a surrogate of the fitness function rather than the mutation function itself~\cite{abboud2001surrogate}.

\section{Mutation Models Framework}

In this paper, we train a machine learning model to imitate the evolutionary process. The model takes a chromosome (in this case, level) as input and outputs a modification to it. The end result is a model that can generate content without the need for an evaluation function. Figure~\ref{fig:system} shows the full system diagram of the mutation models framework, which consists of two parts: the evolution loop and the training loop. The evolution loop is a standard evolutionary algorithm which evolves content by trying to increase fitness. Within the training loop, data is extracted from the evolution loop to train a machine learning model to perform successful mutations (mutations that leads to higher fitness) like the evolutionary process, thus the name \emph{mutation models}. The next two subsections explain both loops in detail.

\subsection{Evolution Loop}\label{sec:evolution}
The evolution loop is a normal, simple evolutionary algorithm. In this work, we use the standard $\mu + \lambda$ evolution strategy without self-adaptation~\cite{beyer2002evolution}. The process as shown in Figure~\ref{fig:system} is simple: it optimizes levels to maximize their fitness. Unlike most evolutionary methods, our method retains the evolution history of every chromosome by recording the process of mutations made to get to its current form. The following steps explain the evolution loop used in this work:
\begin{enumerate}
    \item Generate a random population of levels of size $\mu + \lambda$.
    \item Pick the top $\mu$ levels based on the fitness function to be parents for the next generation.
    \item Generate $\lambda$ new levels by mutating the parent chromosomes ($\mu$ levels).
    \item Repeat the above steps for $G$ generations or until convergence.
\end{enumerate}

We are not restricted to only use a $\mu + \lambda$ evolution strategy algorithm. Other evolutionary algorithms would work well as long as they do not require crossover and allow us to keep track of the evolutionary history of the chromosomes across generations (for example, in a continuous domain it would straightforward to use the high-performing CMA-ES algorithm). However, the selected mutation function will determine how the machine learning algorithm imitates evolution. In this work, the mutations are small, they only change one tile at a time. For example: the system picks a random x,y coordinates (location) in the level and swaps the current tile value with a new tile value or decide not change at all (action). The action does not always need to be random: this could be sampled from a machine learning model trained to perform successful mutations (mutations that improve the fitness function). This will be discussed in the next subsection.

\begin{figure}
    \centering
    \begin{subfigure}[t]{0.47\columnwidth}
         \centering
         \includegraphics[width=0.48\textwidth]{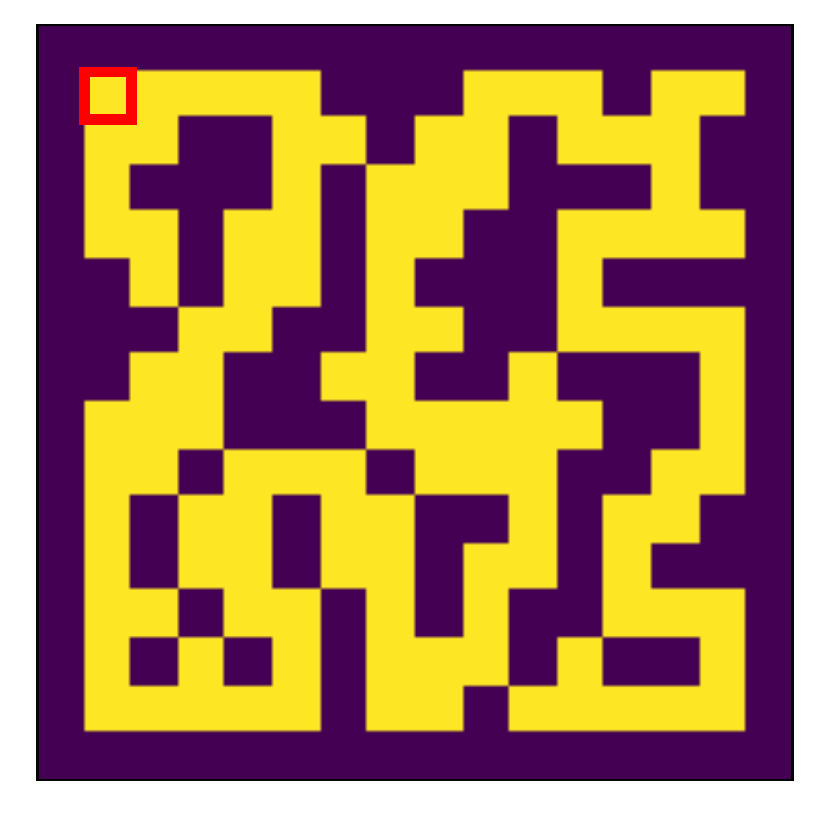}
         \includegraphics[width=0.48\textwidth]{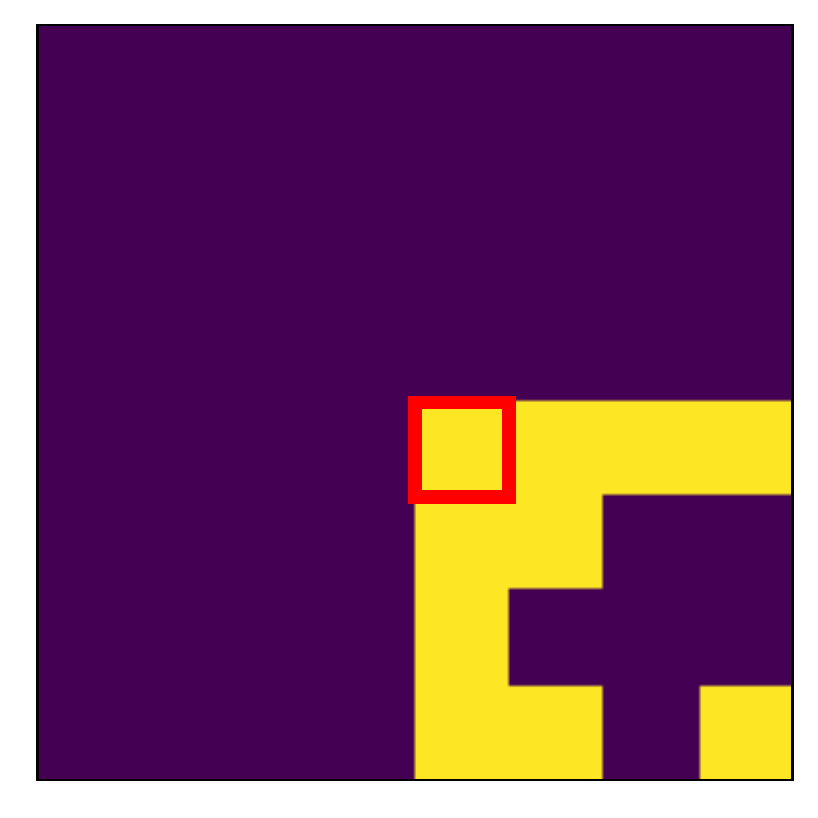}
         \caption{Mutation location (x=0, y=0)}
         \label{fig:cropped_0}
    \end{subfigure}
    \begin{subfigure}[t]{0.47\columnwidth}
         \centering
         \includegraphics[width=0.48\textwidth]{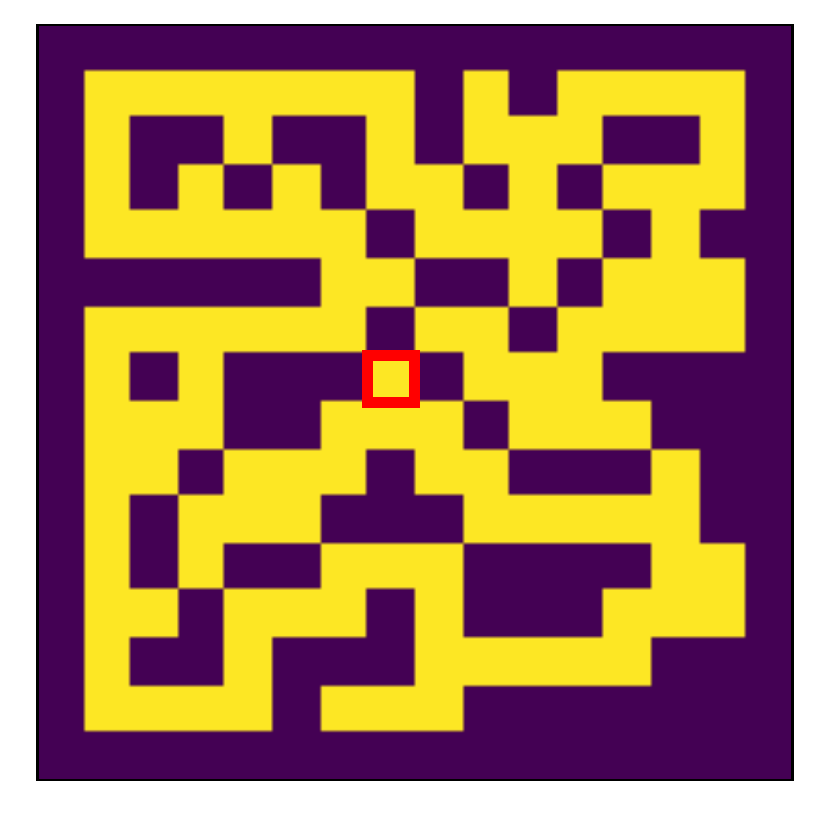}
         \includegraphics[width=0.48\textwidth]{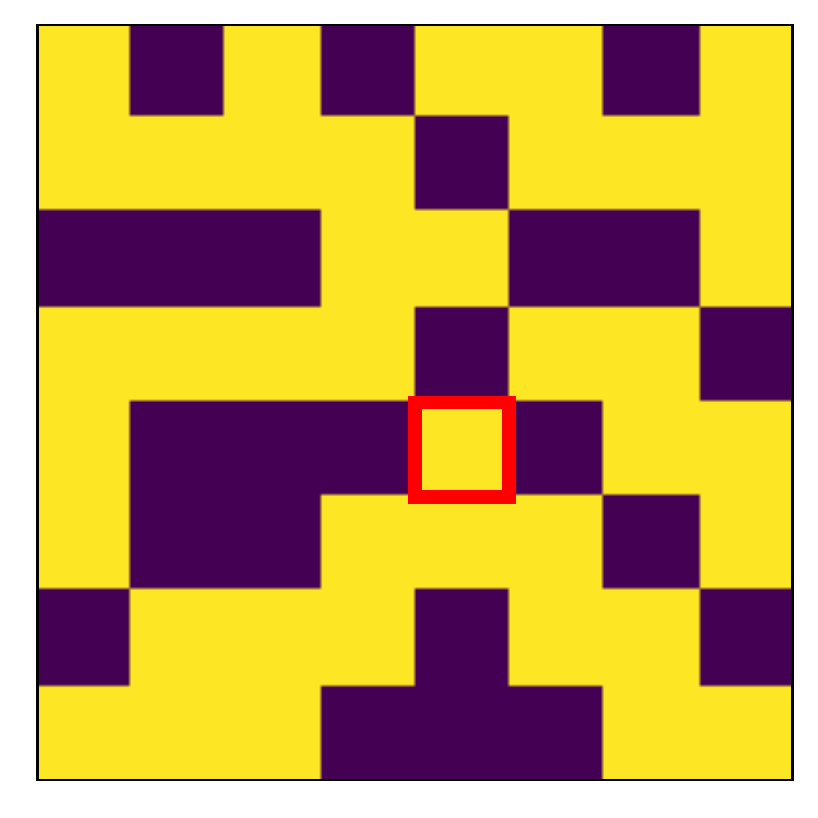}
         \caption{Mutation Location (x=6, y=6)}
         \label{fig:cropped_5}
    \end{subfigure}
    \caption{Examples of transforming a level and a mutation location to a cropped $8x8$ level. This $8x8$ cropped level act as the input for our trained machine learning model.}
    \label{fig:observation}
\end{figure}

\begin{figure*}
    \centering
    \begin{subfigure}[t]{0.3\textwidth}
         \centering
         \includegraphics[width=\textwidth]{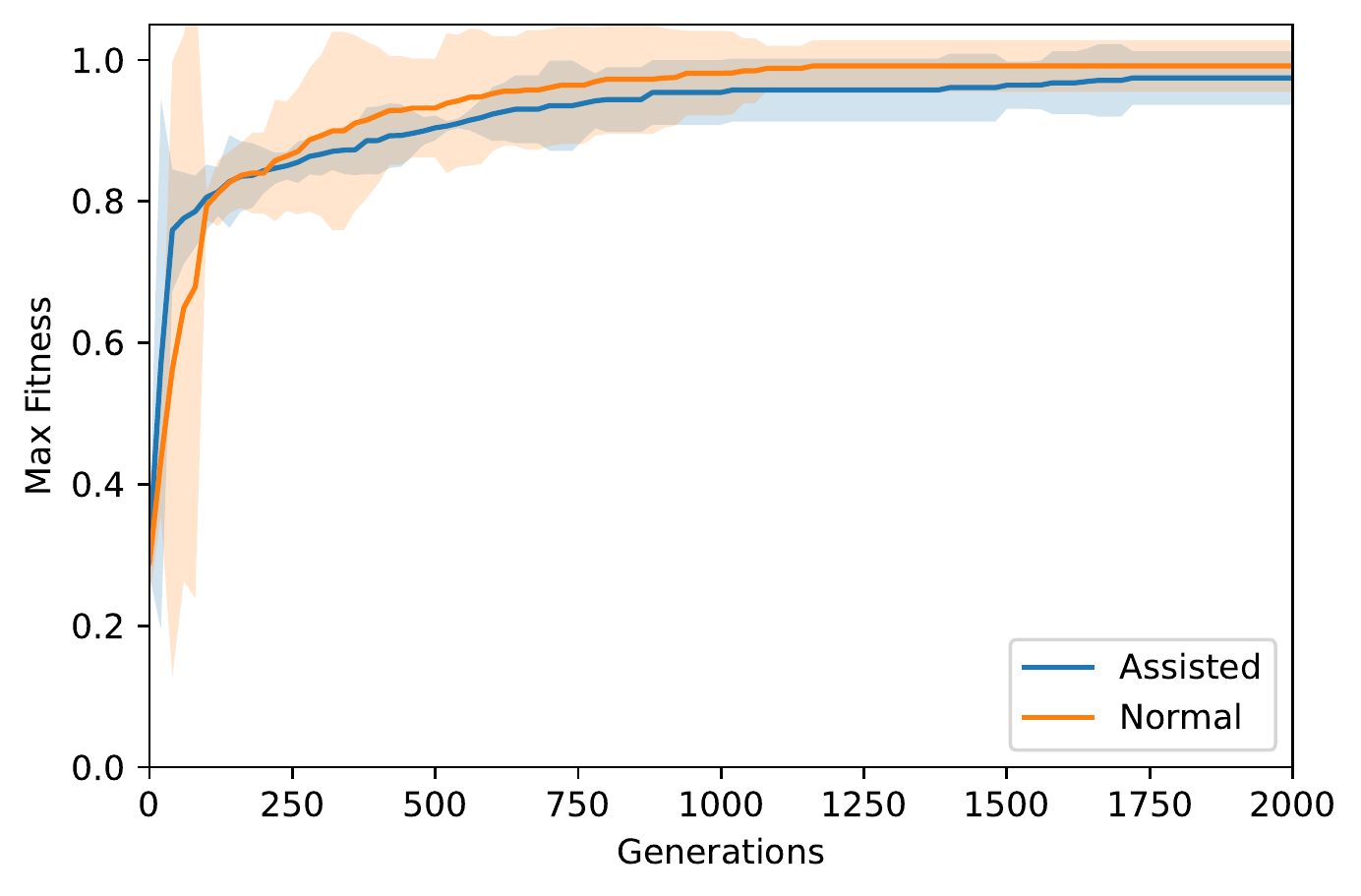}
         \caption{Fitness over time}
         \label{fig:fitness}
    \end{subfigure}
    \begin{subfigure}[t]{0.315\textwidth}
         \centering
         \includegraphics[width=\textwidth]{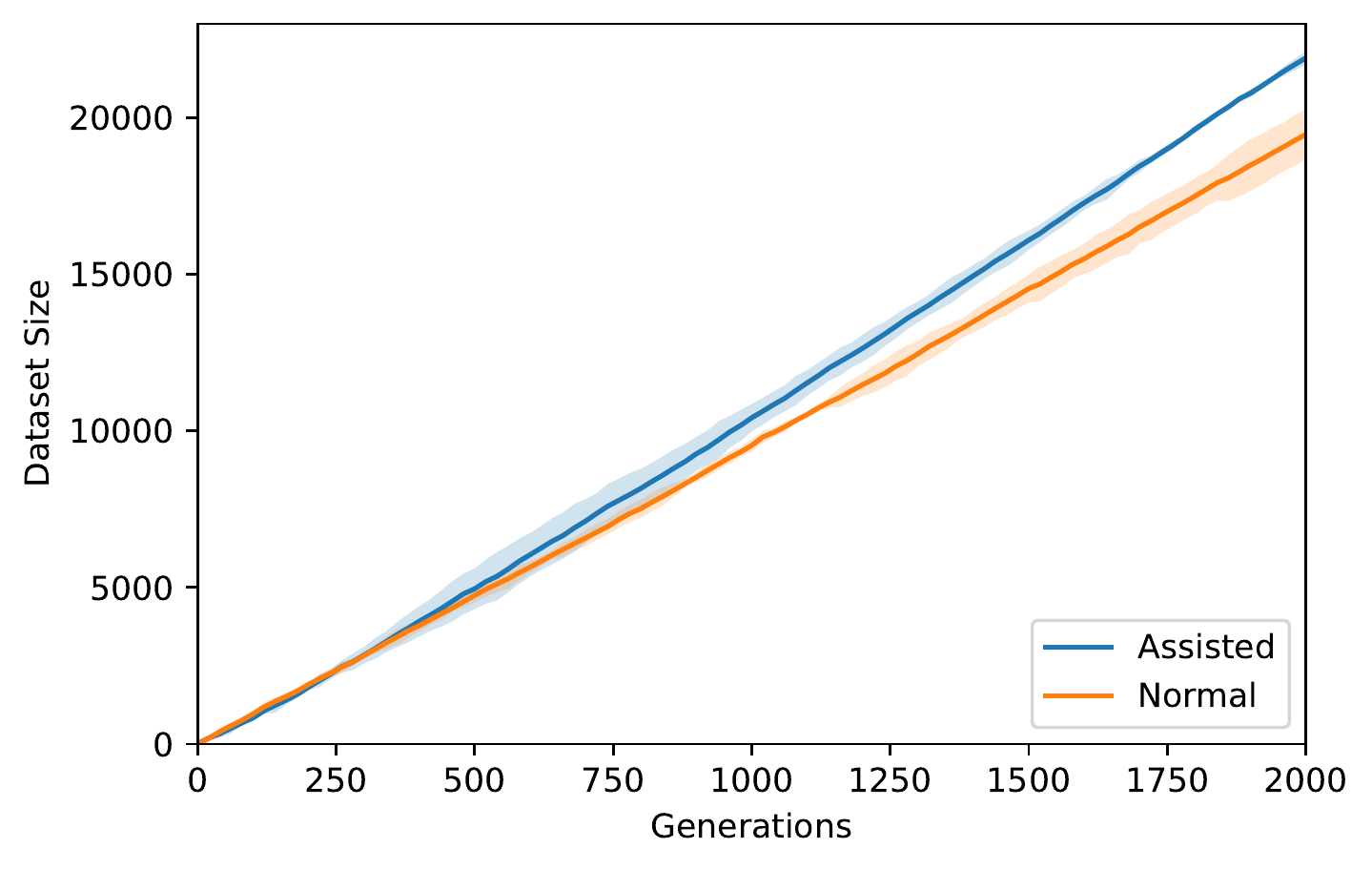}
         \caption{Dataset size over time}
         \label{fig:dataset_size}
    \end{subfigure}
    \begin{subfigure}[t]{0.3\textwidth}
         \centering
         \includegraphics[width=\textwidth]{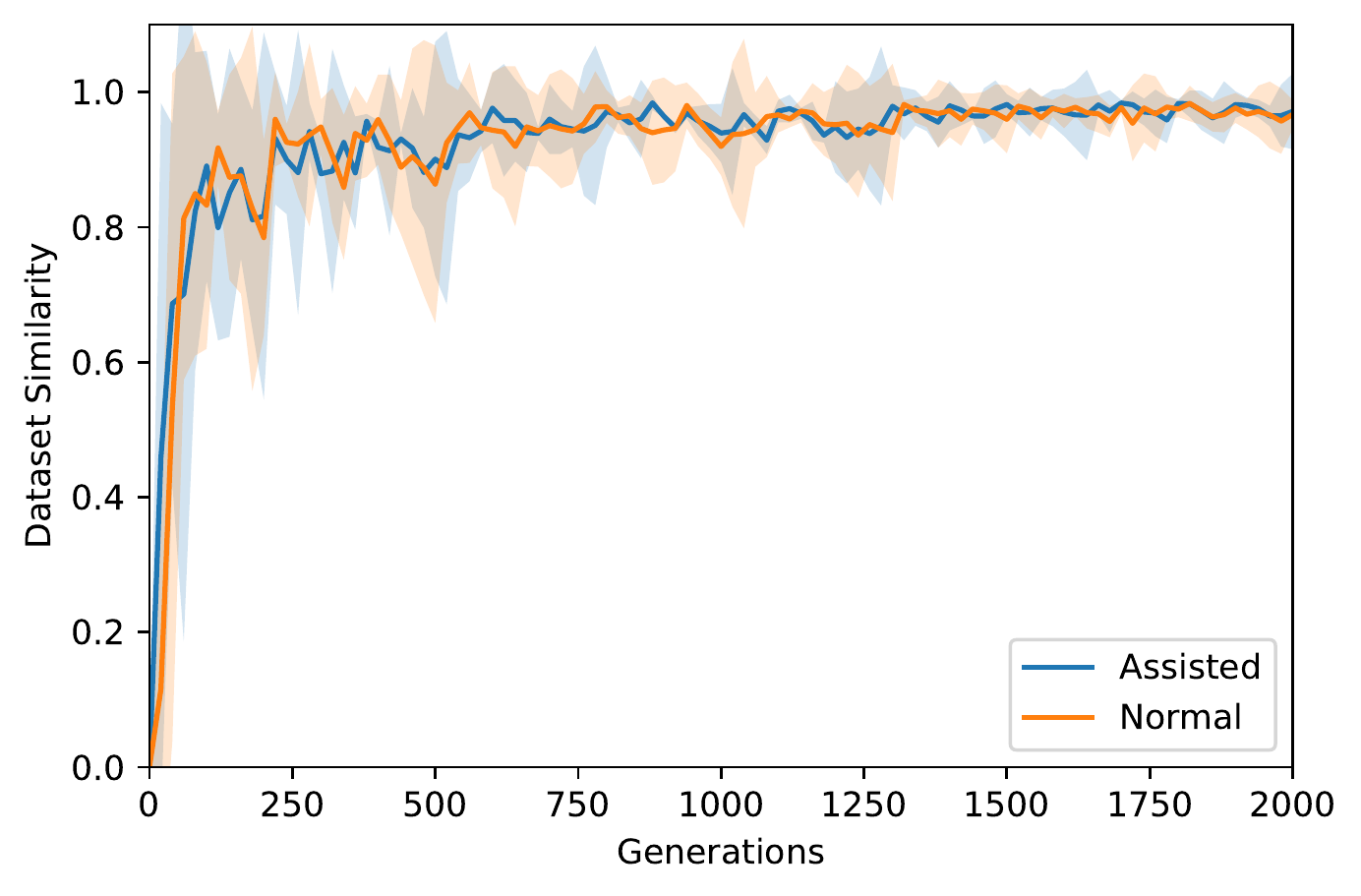}
         \caption{Evolution history similarity over time}
         \label{fig:similarity}
    \end{subfigure}
    \caption{Fitness, dataset size, and dataset similarity of the top 10 chromosomes of each generation over the course of ``Normal'' evolution and ``Assisted'' evolution. The shaded area shows the 95\% confidence interval over 3 runs.}
    \label{fig:evolution_results}
\end{figure*}

\subsection{Training Loop}\label{sec:training}
The training loop is responsible for training a machine learning algorithm to learn the evolutionary process. The idea is to train a model on the successful mutations which cause the top $X$ chromosomes from the population to have a high fitness. Doing so allows the model to learn mutations that are likely to lead to improvement of the level's fitness. The following steps explain the training loop used in this work:
\begin{enumerate}
    \item Pick top $X$ chromosomes from the population.
    \item Extract the trajectory data (evolution history) from these chromosomes to build a dataset of levels and mutations.
    \item Train a machine learning model on the extracted dataset.
\end{enumerate}

The network's training loop can occur at the end of evolution on the top $X$ evolved maps, which we call the ``Normal'' method. It can also occur every $I$ generations, allowing the network to be used as the mutation operator during the evolution loop, which we call the ``Assisted'' method. The ``Assisted'' method is similar in idea to on-policy reinforcement learning: a network creates its own dataset, which it then trains on. In contrast to traditional reinforcement learning, the evolutionary process here is helping filter the training dataset by only keeping the best data points.

In order to build the training dataset, we need to convert the evolution history of a chromosome into inputs and outputs for the machine learning model. We found inspiration in the ``Narrow'' representation in the PCGRL~\cite{khalifa2020pcgrl} environment. In the narrow representation, the input is the mutation location and level state at a certain point in history, while the output is the mutation value (no change or the new tile value). In preliminary experiments, we realized that using the entire level state does not help the machine learning model generalize. We follow the work by Siper et al.~\cite{siper2022path} and Ye et al.~\cite{ye2020rotation} in cropping the level state around the mutation location, as shown in Figure \ref{fig:observation}.

\section{Experiments}
To demonstrate each method's capabilities, we apply them on a simple domain in which they generate 2D maze layouts. We test the two methods of training explained in section~\ref{sec:training}: ``Normal'' and ``Assisted''. The ``Normal'' method trains the model at the end after evolution is done, while ``Assisted'' method retrains the model from scratch every $100$ generations and uses the freshly-trained model as a mutation function in the evolution loop. All our experiments are repeated $3$ times (the full evolution loop including training a neural network either at the end of evolution in the case of ``Normal'' or during evolution in the case of ``Assisted''), and we show the average max fitness, dataset size, and dataset similarity across these runs and the 95\% confidence interval. After evolution is done, the final trained models from each method are tested for their capabilities on generating content without the need for a fitness function.

\subsection{Domain}
We use a 2D maze layout of size $14x14$ (excluding the borders) as our proof-of-concept test-bed. The goal is to generate a 2D layout of solid and empty tiles such that all the empty tiles are accessible from any other empty tile and the longest path in that layout is maximized. Figure~\ref{fig:evolution_maps} shows the top 4 evolved maps by the evolutionary algorithm showcasing full connectivity and long winding paths.

\subsection{Evolution Loop}
As discussed in section~\ref{sec:evolution}, we are using $\mu + \lambda$ evolution strategy with $\mu = \lambda = 50$. The starting levels are initialized using a uniform distribution with $50\%$ probability for each tile to be either empty or solid. The mutation operator picks a random location ($x$ and $y$ position) and uses either a random mutation value for the ``Normal'' method, or samples the value from the trained machine learning model for the ``Assisted'' method. To guarantee diversity in the ``Assisted'' method, there is a $25\%$ chance to sample a random action instead from the trained model. We run the evolution process for $2000$ generations with a cascading fitness function which tries to satisfy the connectivity constraint first then maximizes path length as shown in equation~\ref{eq:fitness}:
\begin{equation}\label{eq:fitness}
    f = \begin{cases}
          0.5 \cdot (1 - \frac{n}{20}) & \text{if } n > 1 \text{ and } n < 20 \\
          0.5 + 0.5 \cdot \frac{p}{98} & \text{if } n == 1\\
          0 & otherwise\\
    \end{cases}
\end{equation}
where $n$ is the number of regions in the current level and $p$ is the length of the longest path length in the map.

\subsection{Training Loop}

The training loop is responsible about two things, creating a training dataset for the model and training the model. For the dataset generation, we pick the top $10$ chromosomes from the population and extract their evolution history. The evolution history consists of the current level, the mutation location, and the mutation action at every generation in a chromosome's history. We transform the level and mutation location into an single image by adopting the narrow representation from the PCGRL~\cite{khalifa2020pcgrl} framework. We crop the level around the mutation location to be of size $8x8$ as shown in Figure~\ref{fig:observation}. This cropped level is used as the input for the machine learning model. The output of the model is the mutation action that corresponds to the cropped level. In this domain, there are 3 different mutation actions: ``No Change'', ``Change to Empty'', and ``Change to Solid''.

The machine learning model is a small convolutional neural network (224,771 parameters) with a similar architecture to the Atari Deep-Q network~\cite{mnih2013playing}. The network consists of $3$ convolutional layers (with $3x3$ filters and kernel size of $32$, $64$, and $128$ respectively) followed by $2$ fully connected layers (with $256$ and $3$ neurons respectively). We use 2 max pooling layers, each size 2, after the first and second convolution layers to decrease the input space size. All convolutional layers use same padding to make sure the size of the input remains constant. All activation functions are Relu except for the last layer, where we use softmax. The neural network is trained using an Adam optimizer with learning rate of $10^{-4}$, a minibatch size of $32$, and categorical cross entropy loss. For the ``Normal'' method, we train a new network for $2$, $4$, and $8$ epochs to explore how different epochs impact network results. For the ``Assisted'' method, we only train the network for $2$ epochs due to the computation costs. The ``Assisted'' method network is retrained from scratch (with random weights and a reset learning rate) every $100$ generations. This decision was decided as preliminary experiments showed that models trained continuously had a higher chance to overfit on earlier data.

\subsection{Model Inference}

During the evolution/training step, both the ``Normal'' and ``Assisted'' methods utilize a fitness function. The evolutionary algorithm creates a dataset which is used to train a neural network to imitate the evolutionary process. After the model is trained (during \emph{inference}), the trained model does \textbf{not} use any fitness function in \textbf{either} method. The inference step is the same regardless of which evolution/training procedure is used.
We use the final trained network from each experiment to generate $100$ levels starting from noise ($50\%$ solid and $50\%$ empty levels). The network actions are sampled using the probability distribution from the last softmax layer. The trained networks are run sequentially like a scanline over the entire map multiple times, as proposed in~\cite{khalifa2020pcgrl}. The networks stop iterating if they reach one of two cases: success (all the empty tiles are fully connected) or failure (iterated on the whole map for $196$ times). We record the success rate and the iteration number that the model requires to reach a successful level. We also record the path length of each of the successful generated levels with the number of empty tiles, which we use to calculate the diversity of the generated content. This diversity is shown both as an expressive range analysis as well as a percentage of generated levels having different combinations of path lengths and empty tiles.

\section{Results}
In this section, we explore the results from our system. We start showing the evolution results in Section \ref{sec:results-evolution}. These results are collected from the evolutionary generator which uses an evaluation function to create levels. We analyze the effect of using the ``Normal'' vs ``Assisted'' methods on the evolution loop. In contrast, Section \ref{sec:results-model} shows the results of the model inference. The results are collected from the final trained models, which do not use a fitness function during inference.

\subsection{Evolution Results}\label{sec:results-evolution}

\begin{figure}
    \centering
    \begin{subfigure}[t]{0.97\linewidth}
         \centering
         \includegraphics[width=0.24\textwidth]{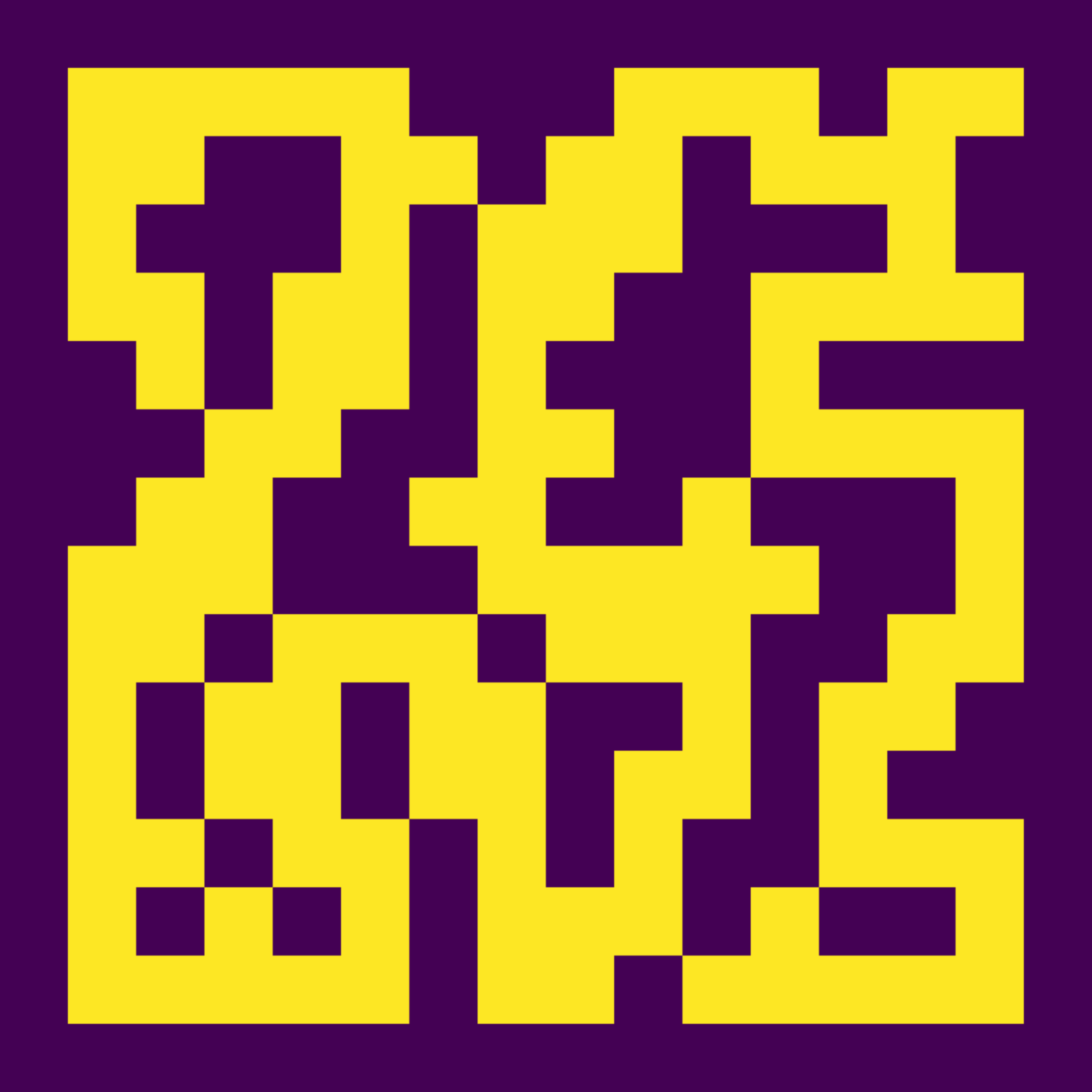}
         \includegraphics[width=0.24\textwidth]{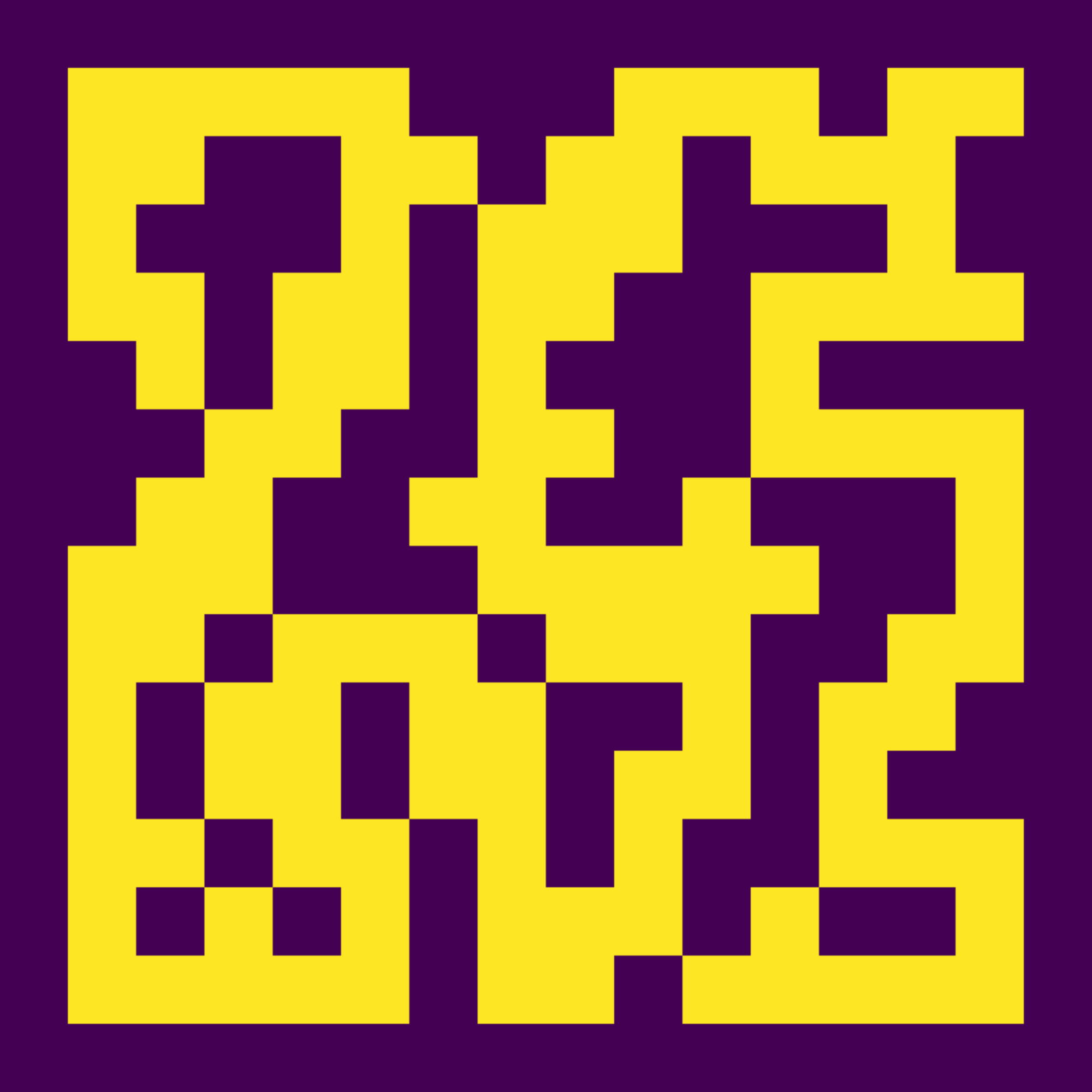}
         \includegraphics[width=0.24\textwidth]{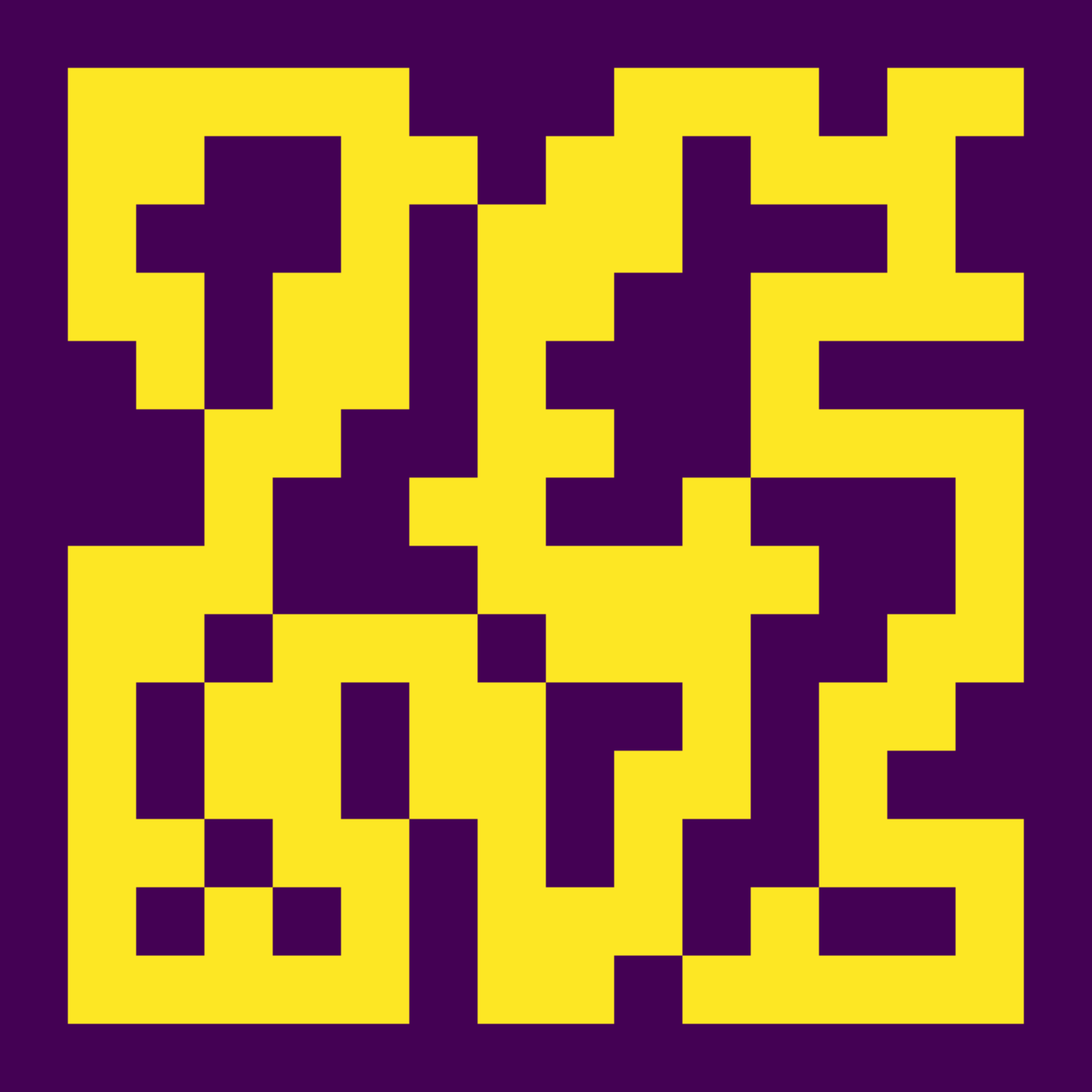}
         \includegraphics[width=0.24\textwidth]{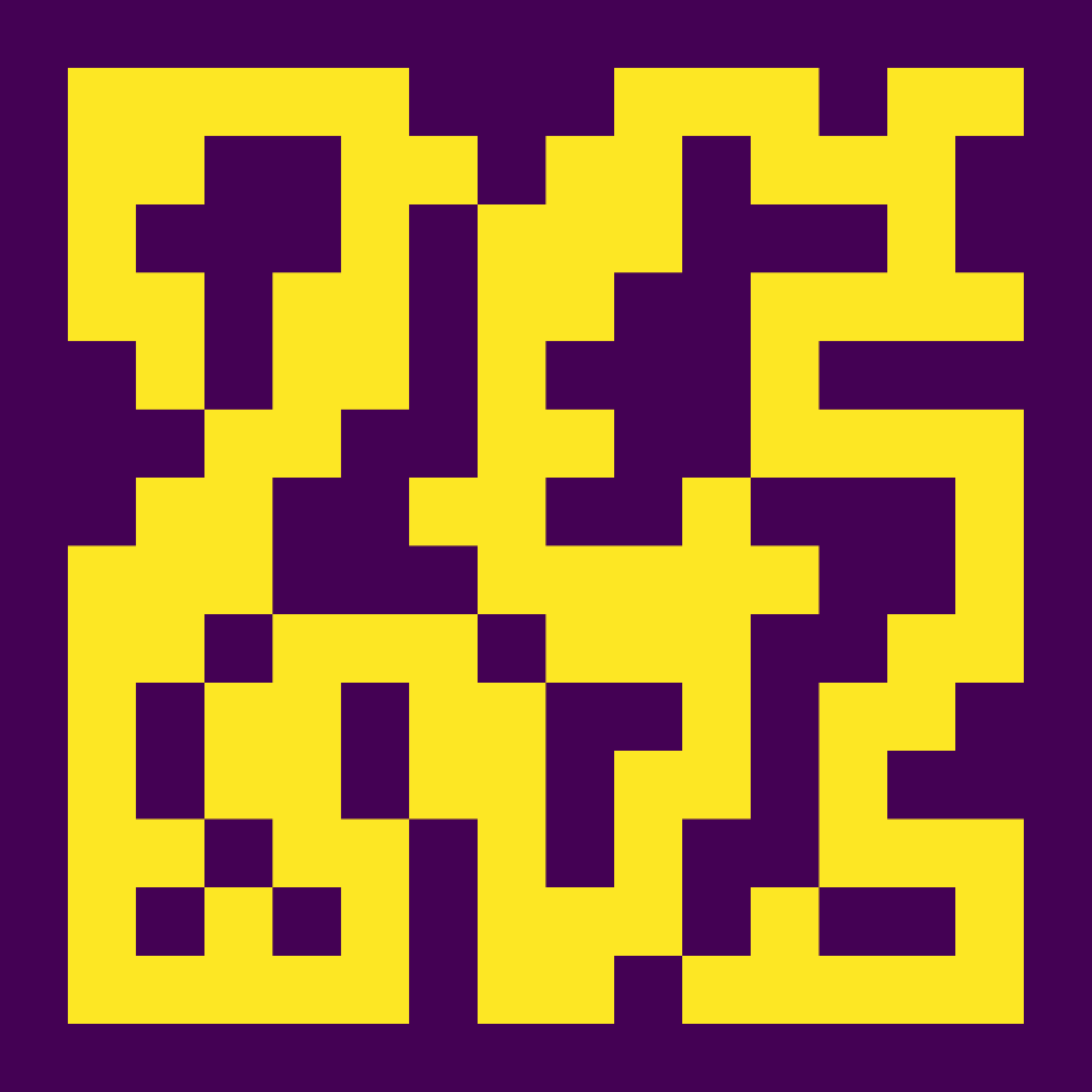}
         \caption{Top 4 generated maps from ``Assisted'' Evolution}
         \label{fig:goal_assisted}
    \end{subfigure}
    \begin{subfigure}[t]{0.97\linewidth}
         \centering
         \includegraphics[width=0.24\textwidth]{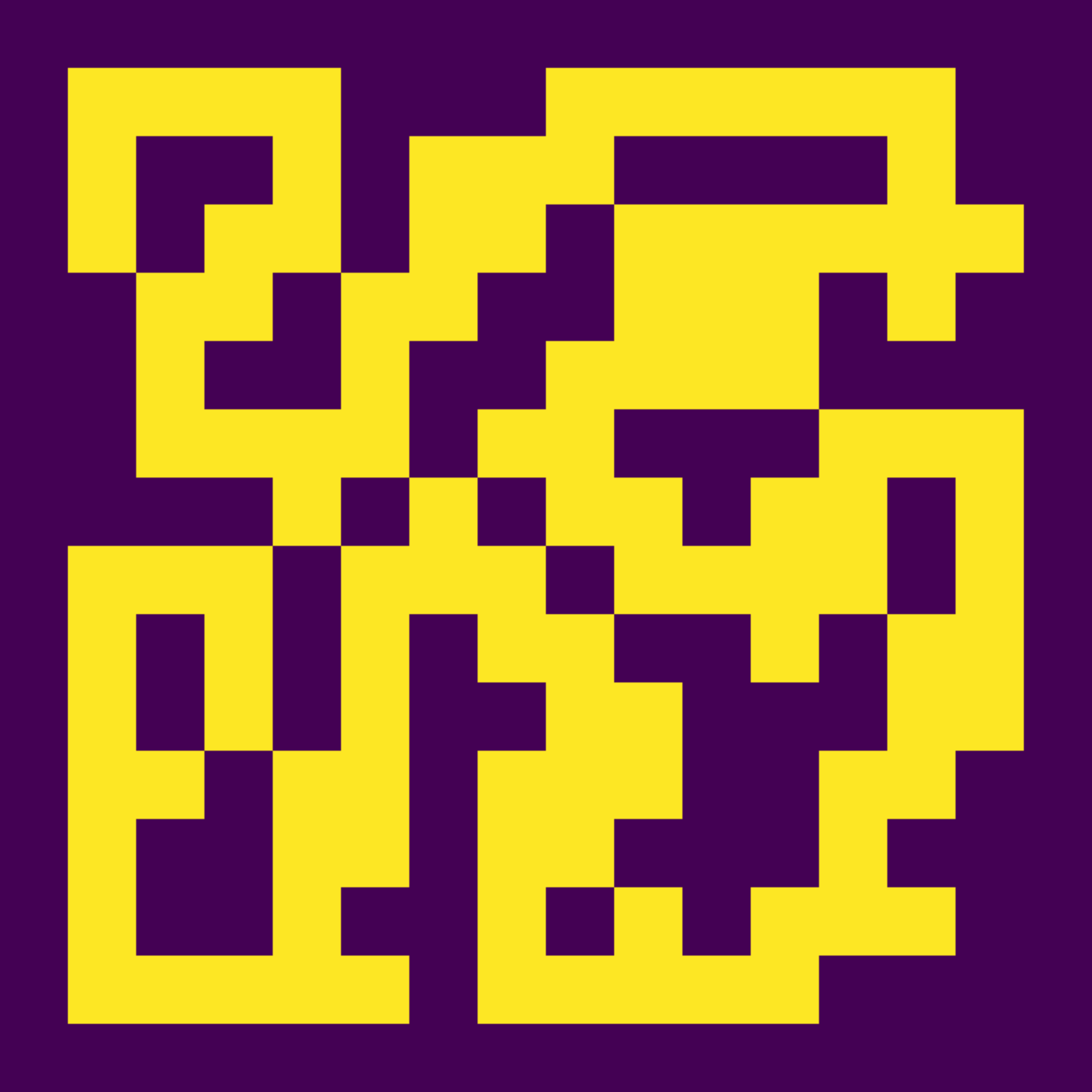}
         \includegraphics[width=0.24\textwidth]{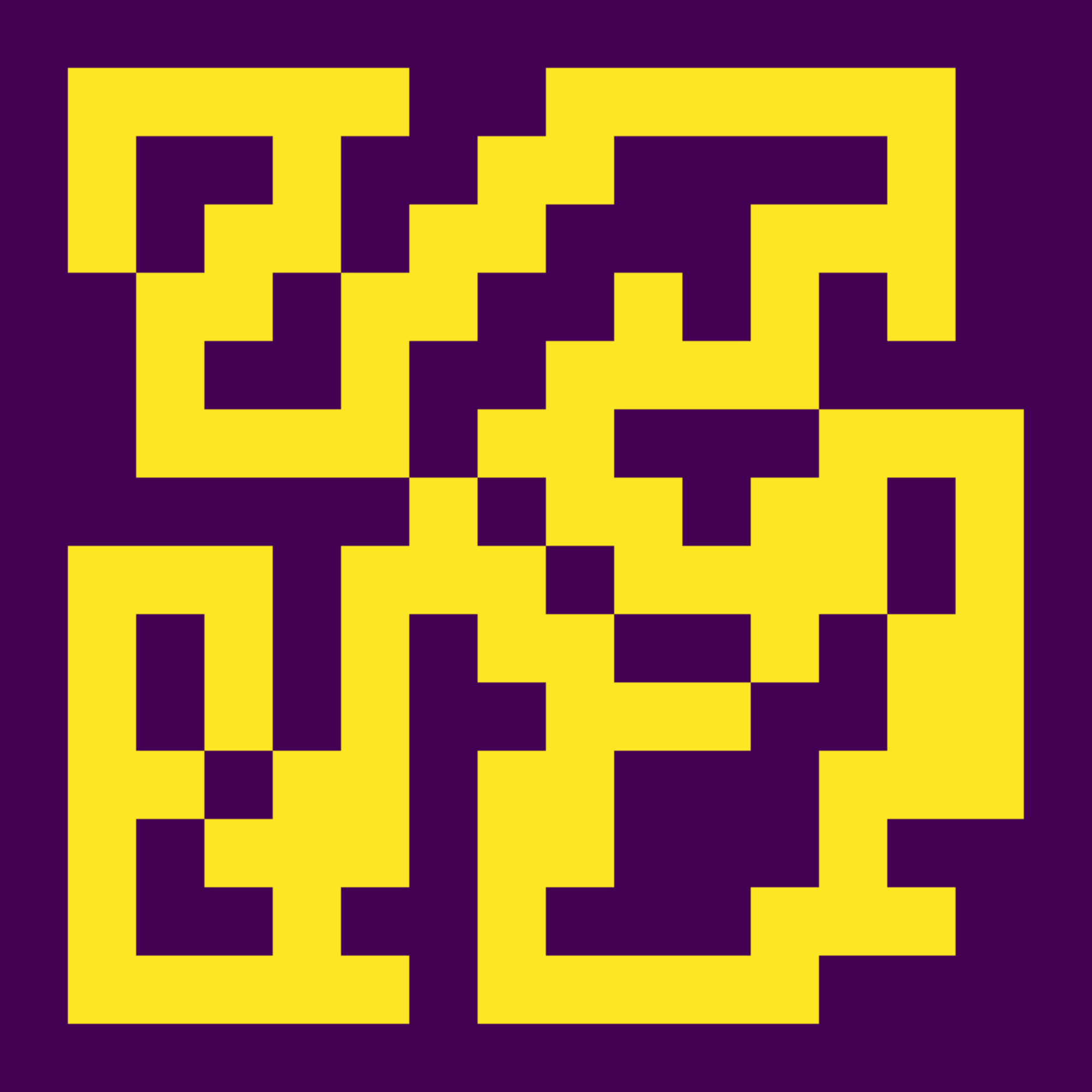}
         \includegraphics[width=0.24\textwidth]{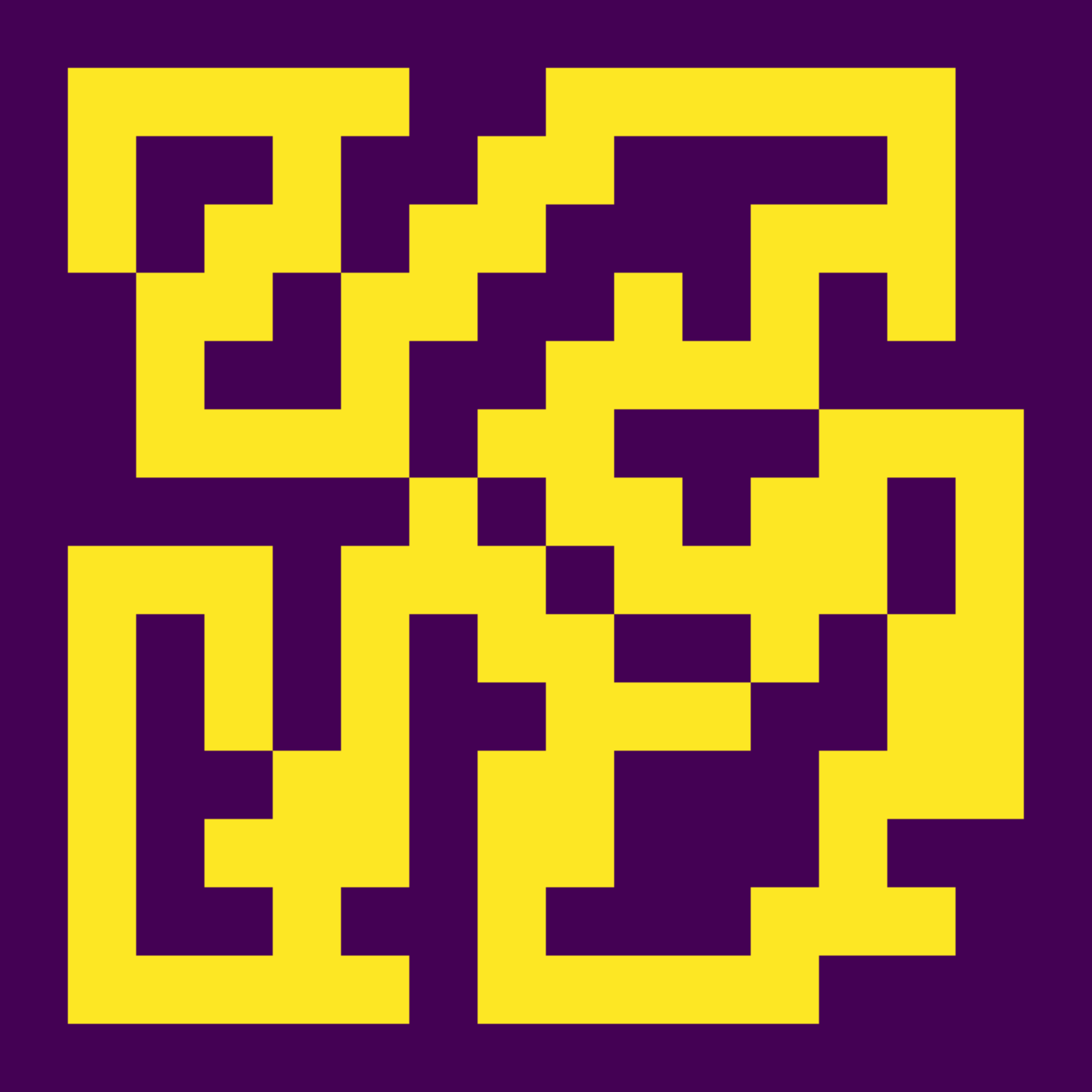}
         \includegraphics[width=0.24\textwidth]{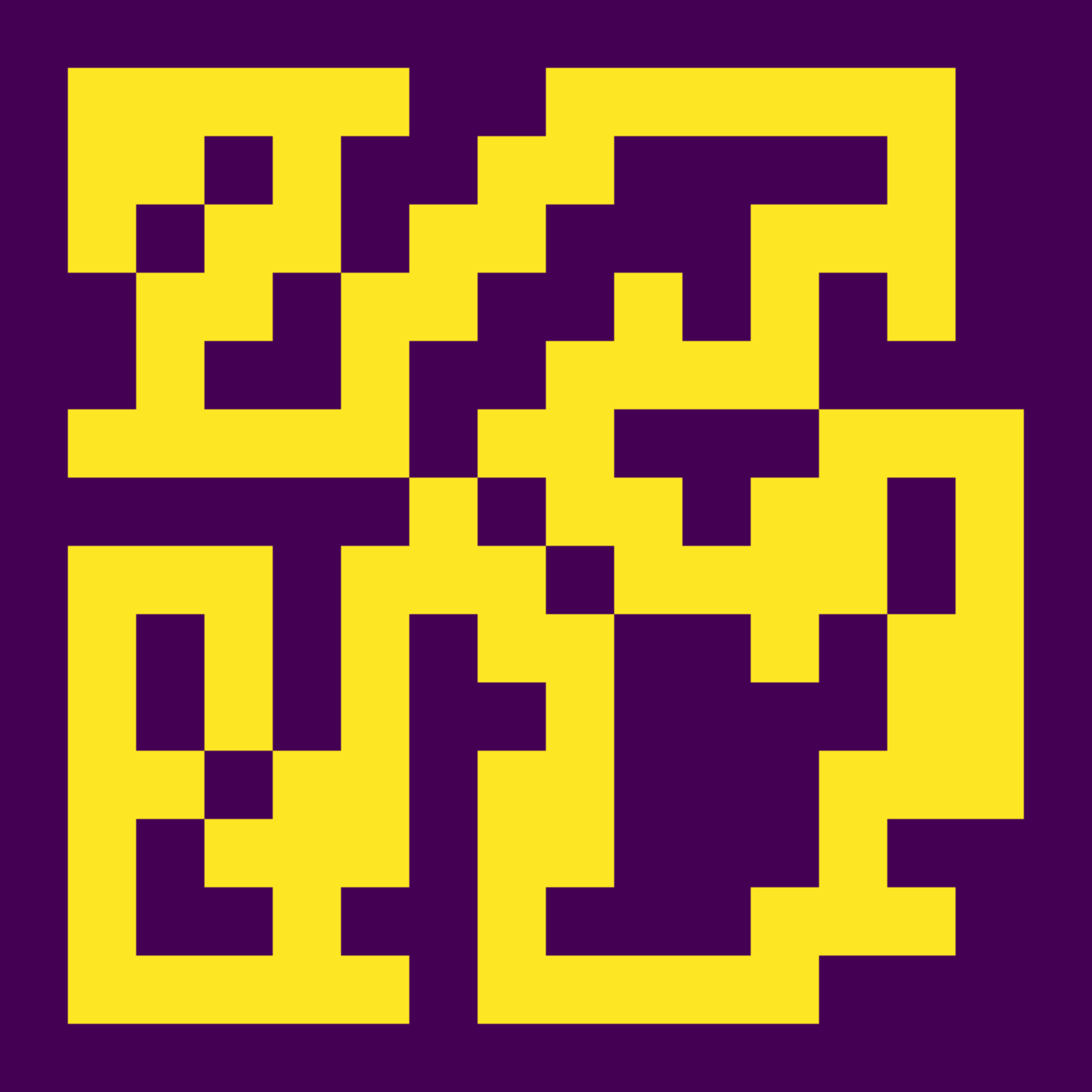}
         \caption{Top 4 generated maps from ``Normal'' Evolution}
         \label{fig:goal_final2}
    \end{subfigure}
    \caption{Top four maps at generation 2000, from two different evolutionary runs (``Assisted'' evolution and ``Normal'' evolution). All the four maps look almost identical with few tile differences as evolutionary strategy tends to converge when most chromosomes in the population become highly similar to each other.}
    \label{fig:evolution_maps}
\end{figure}

Figure~\ref{fig:evolution_results} displays different evolution metrics across the $2000$ generations. The usage of ``Assisted'' method does not change the fitness performance of the evolutionary algorithm with respect to the ``Normal'' method. The only noticeable difference is the size of the extracted dataset (Figure~\ref{fig:dataset_size}) where the ``Assisted'' method manages to create a slightly larger dataset than the ``Normal'' method. This could be an effect of the network refusing to change some chromosomes early (using the``No Change'' action) when the mutation location is not good enough. This would lead to these chromosomes having a longer history than usual as they survive for multiple generations. 

The last metric we observe is the similarities between the evolution histories of the top $10$ chromosomes. We notice that the top $10$ begin with very different trajectories but end up having almost identical trajectories with only slight differences (Figure~\ref{fig:similarity}). This was expected as evolutionary strategy tends to converge when most chromosomes in the population become highly similar to each other as shown in Figure~\ref{fig:evolution_maps}. In future work, this might be solved by using a more complex evolutionary algorithm or a quality diversity algorithm such as MAP-Elites~\cite{mouret2015illuminating} which could ensure diverse trajectories. The similarity metric that we use is based on the Ratcliff-Obershelp algorithm~\cite{black2022ratcliff}.

\subsection{Model Inference Results}\label{sec:results-model}

\begin{table}
    \centering
    \resizebox{\columnwidth}{!}{%
    \begin{tabular}{|l||c|c|c|}
    \hline
        Method & Success & Diversity & Iterations\\
        \hline
        \hline
        Assisted-2 & \textbf{99.67\% ± 0.49\%} & \textbf{86.83\% ± 3.8\%} & \textbf{18.21 ± 18.57} \\
        \hline
        Normal-2 & 30.17\% ± 32.7\% & 28.5\% ± 30.62\% & 61.7 ± 47.22 \\
        Normal-4 & 66.83\% ± 49.22\% & 59.33\% ± 43.69\% & 23.55 ± 48.59 \\
        Normal-8 & 65.17\% ± 48.37\% & 60\% ± 44.59\% & 31.78 ± 23.84 \\
        \hline
    \end{tabular}%
    }
    \caption{The average and 95\% confidence interval of 100 generated levels from 3 runs for success rate, diversity, and number of iterations by the final trained network from ``Assisted'' and ``Normal'' evolution. The ``-\#'' after the name of an experiment designates the number of epochs the final network used for training.}
    \label{tab:play_diverse_time}
    \vspace{-5mm}
\end{table}

In this section, we compare the results between the final trained network from the ``Assisted'' method using $2$ epochs ($Assisted-2$) and the final trained networks from the ``Normal'' method using $2$, $4$, and $8$ epochs ($Normal-2$, $Normal-4$, and $Normal-8$ respectively). Table~\ref{tab:play_diverse_time} displays the success rate and diversity of the generated levels as well as the number of iterations each model needs to reach success (which we will refer from now on as ``iterations''). The diversity and iterations are only calculated on the successfully generated levels, and not for failures.

\begin{figure}
    \centering
    \begin{subfigure}[t]{0.97\linewidth}
         \centering
         \includegraphics[width=0.24\textwidth]{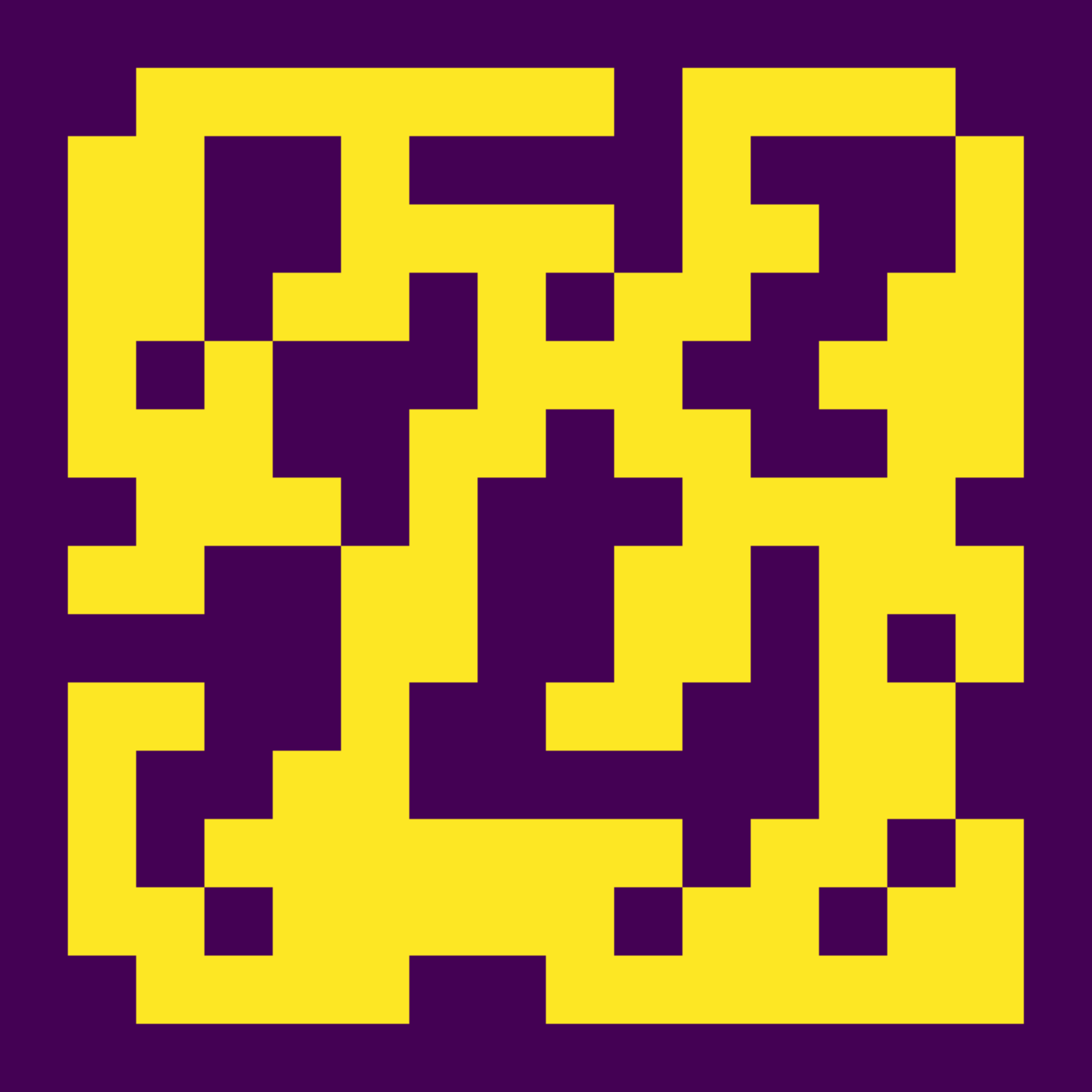}
         \includegraphics[width=0.24\textwidth]{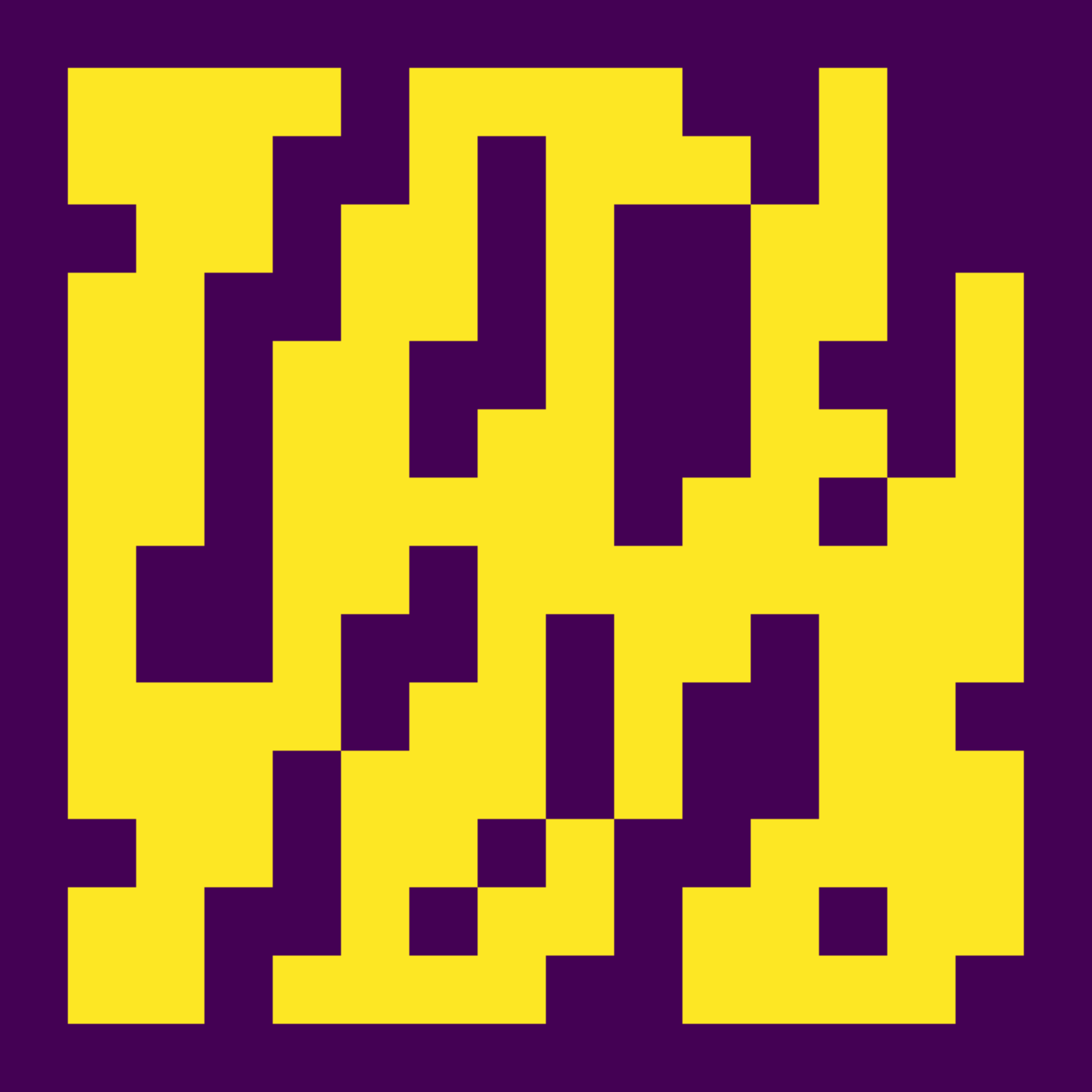}
         \includegraphics[width=0.24\textwidth]{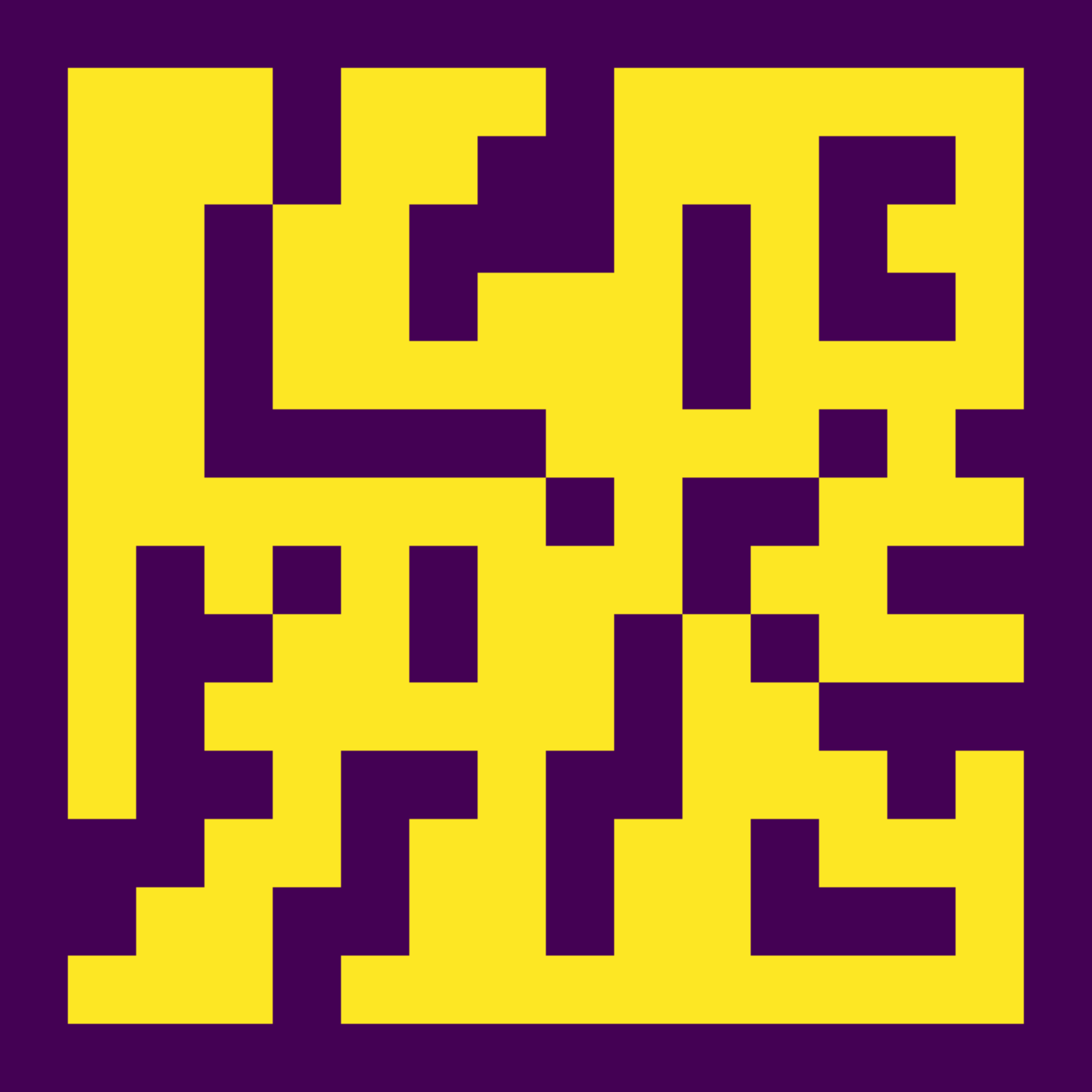}
         \includegraphics[width=0.24\textwidth]{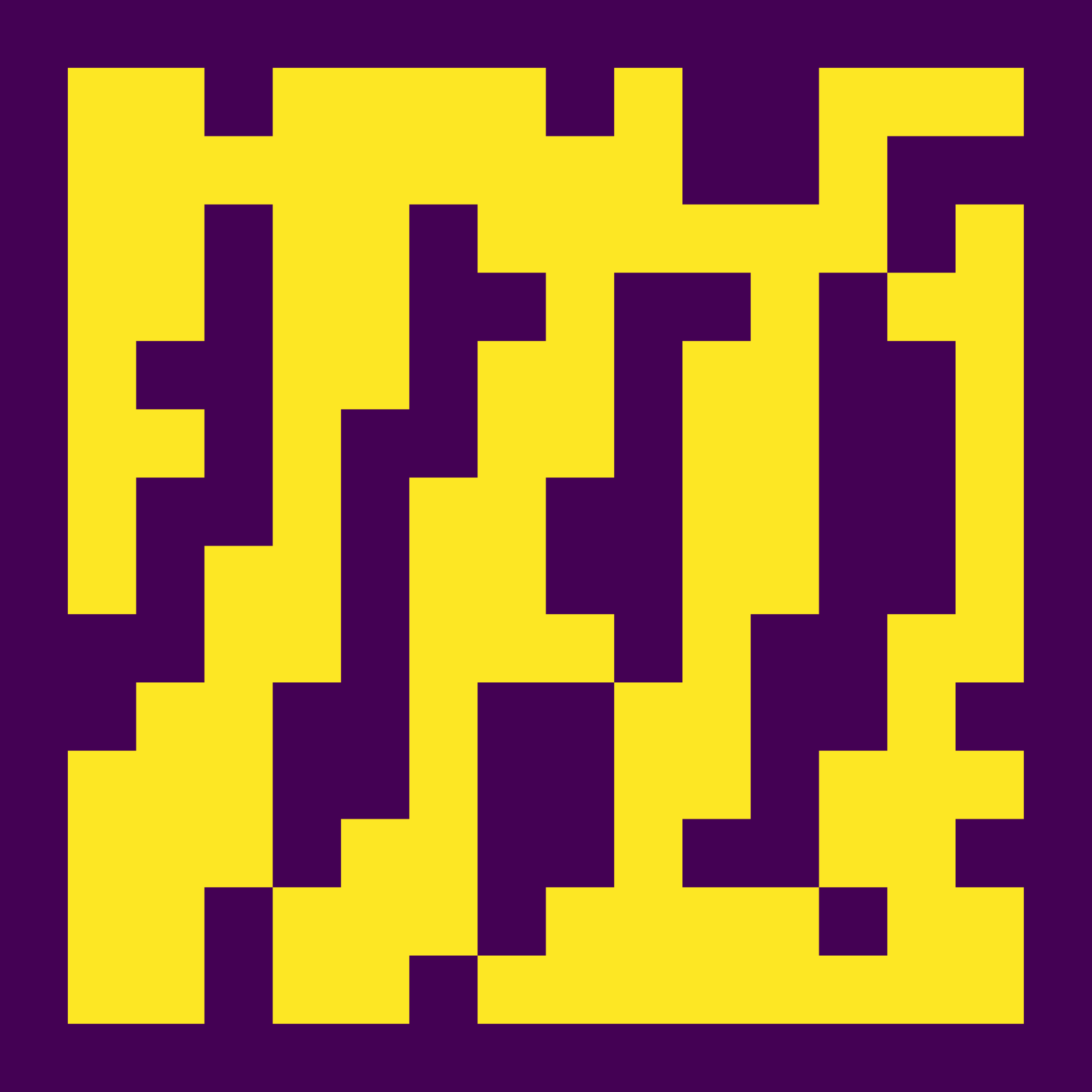}
         \caption{Assisted-2}
         \label{fig:result_assisted}
    \end{subfigure}
    \begin{subfigure}[t]{0.97\linewidth}
         \centering
         \includegraphics[width=0.24\textwidth]{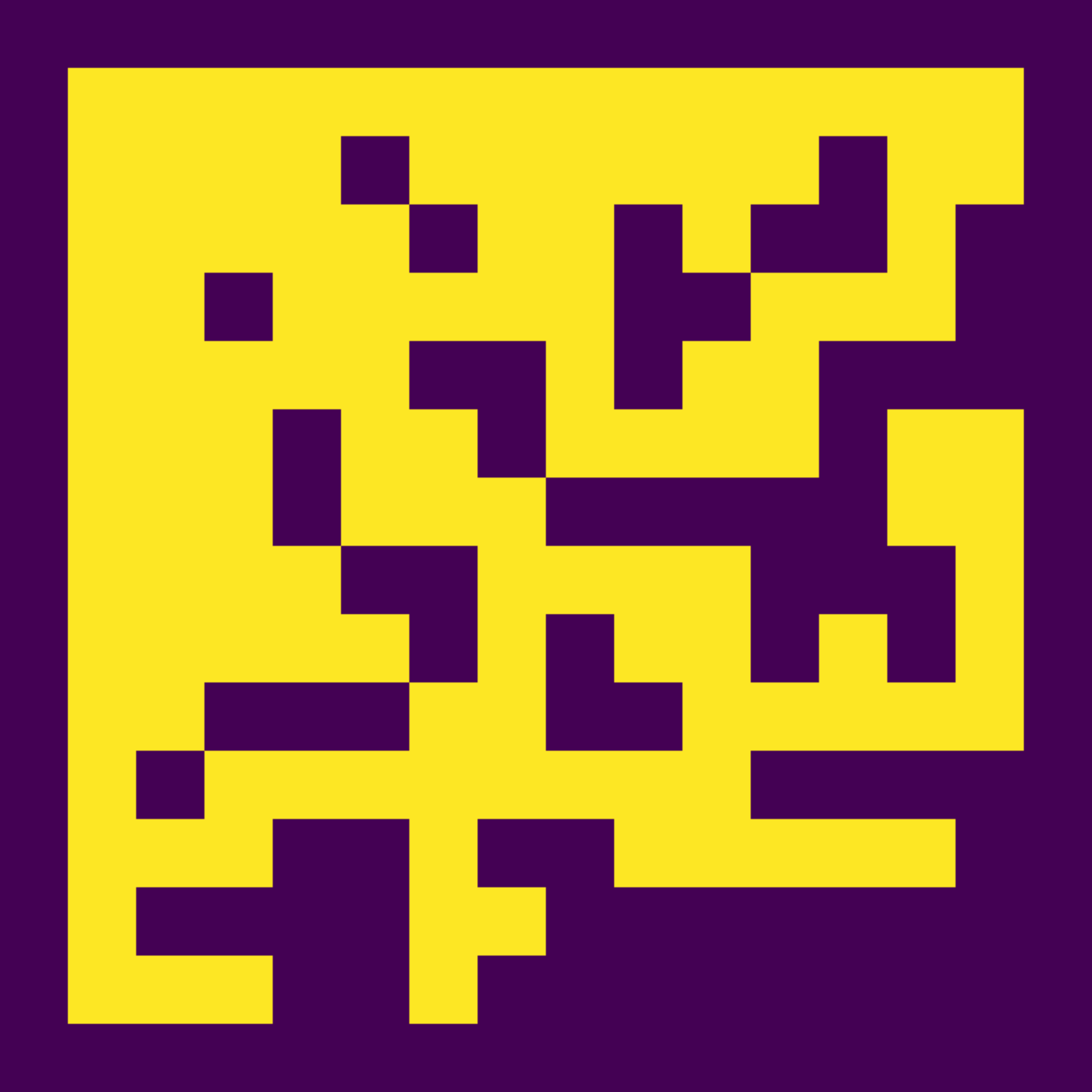}
         \includegraphics[width=0.24\textwidth]{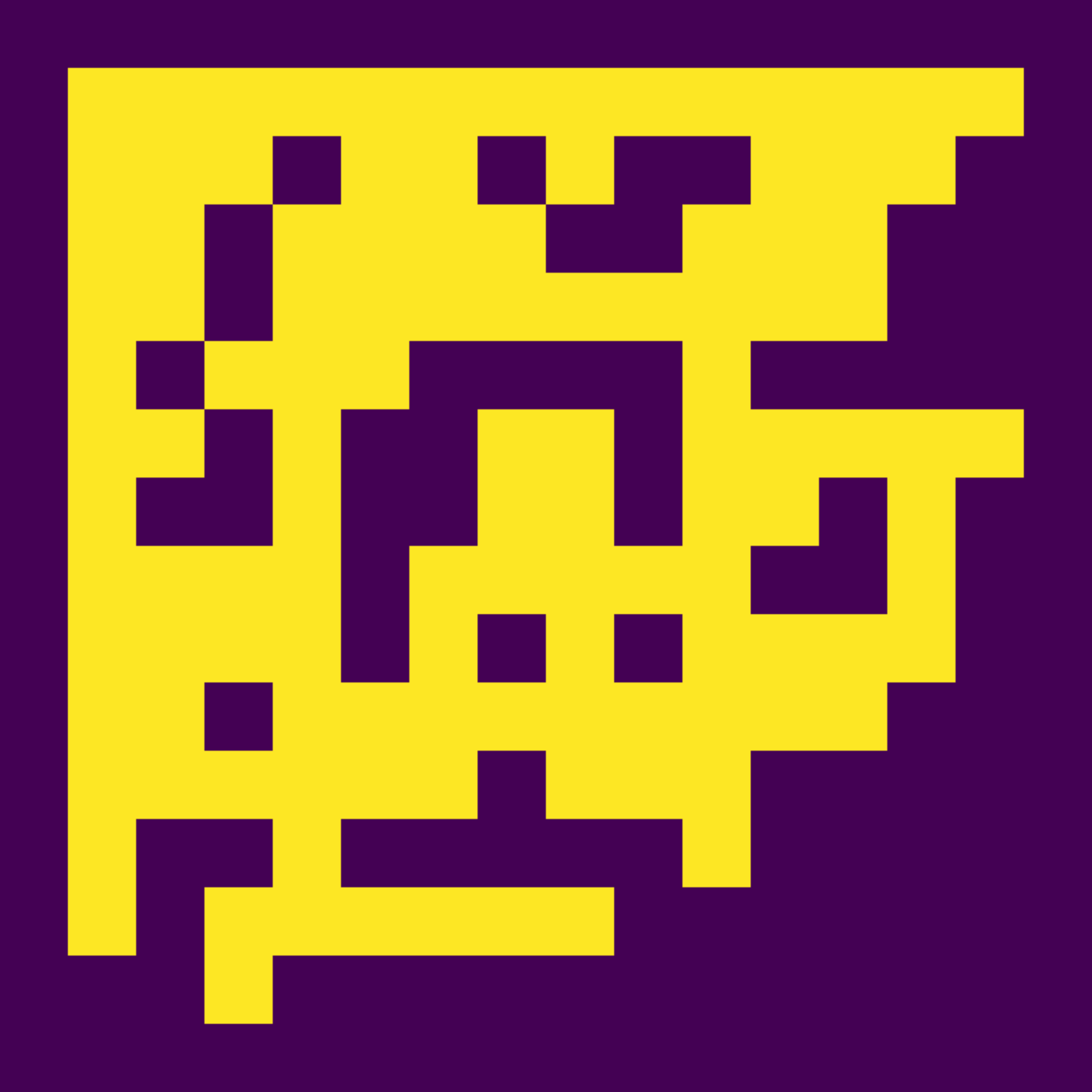}
         \includegraphics[width=0.24\textwidth]{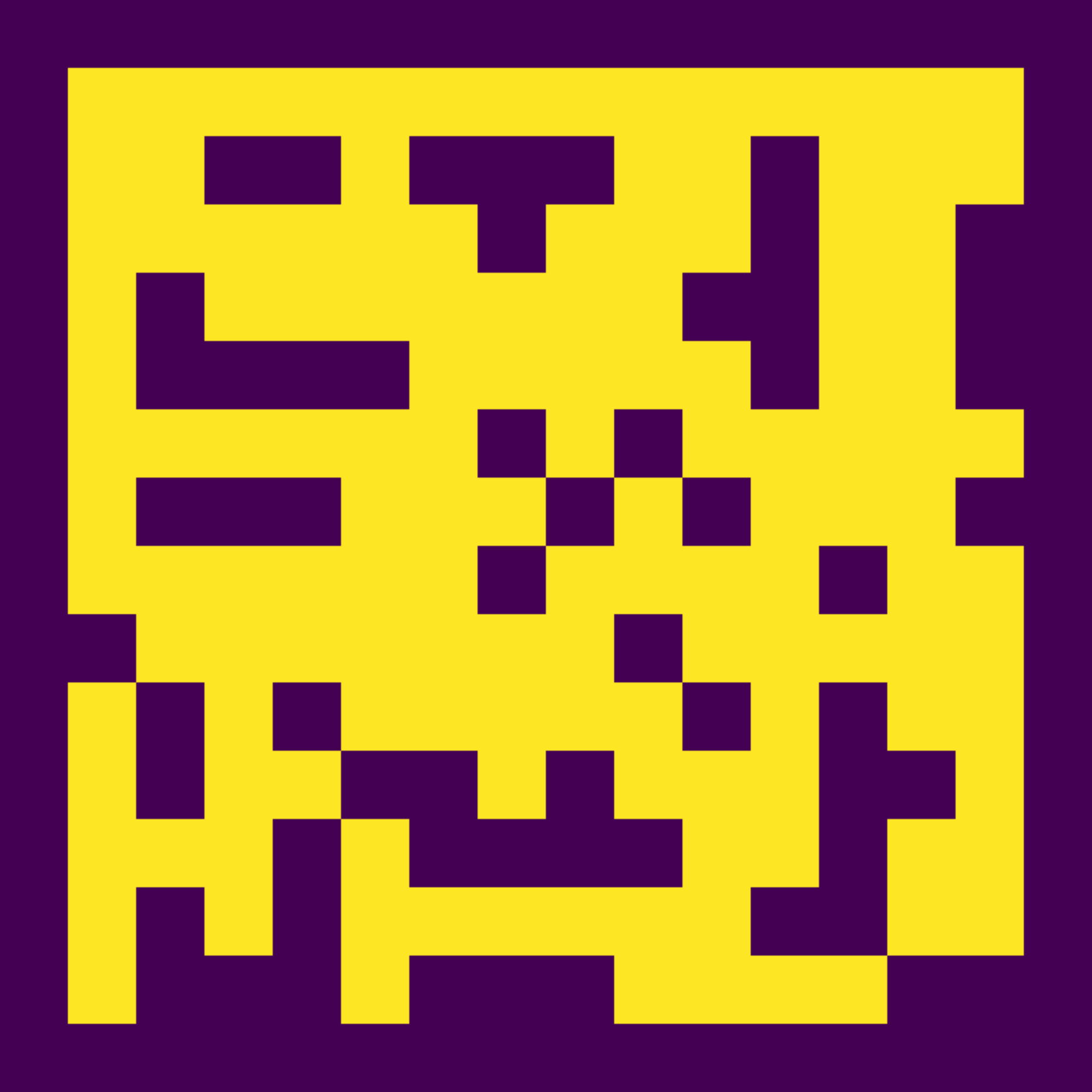}
         \includegraphics[width=0.24\textwidth]{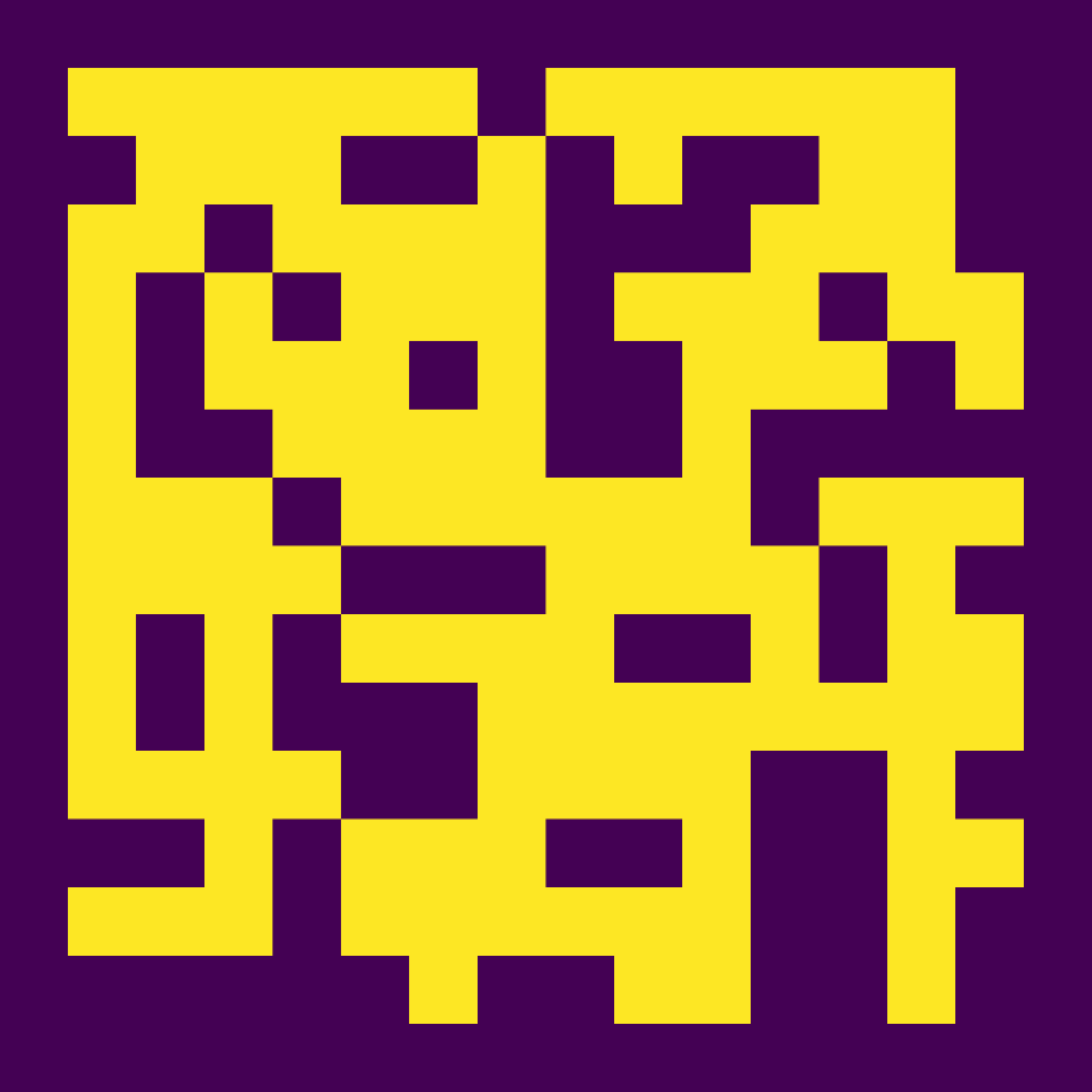}
         \caption{Normal-2}
         \label{fig:result_final2}
    \end{subfigure}
    \begin{subfigure}[t]{0.97\linewidth}
         \centering
         \includegraphics[width=0.24\textwidth]{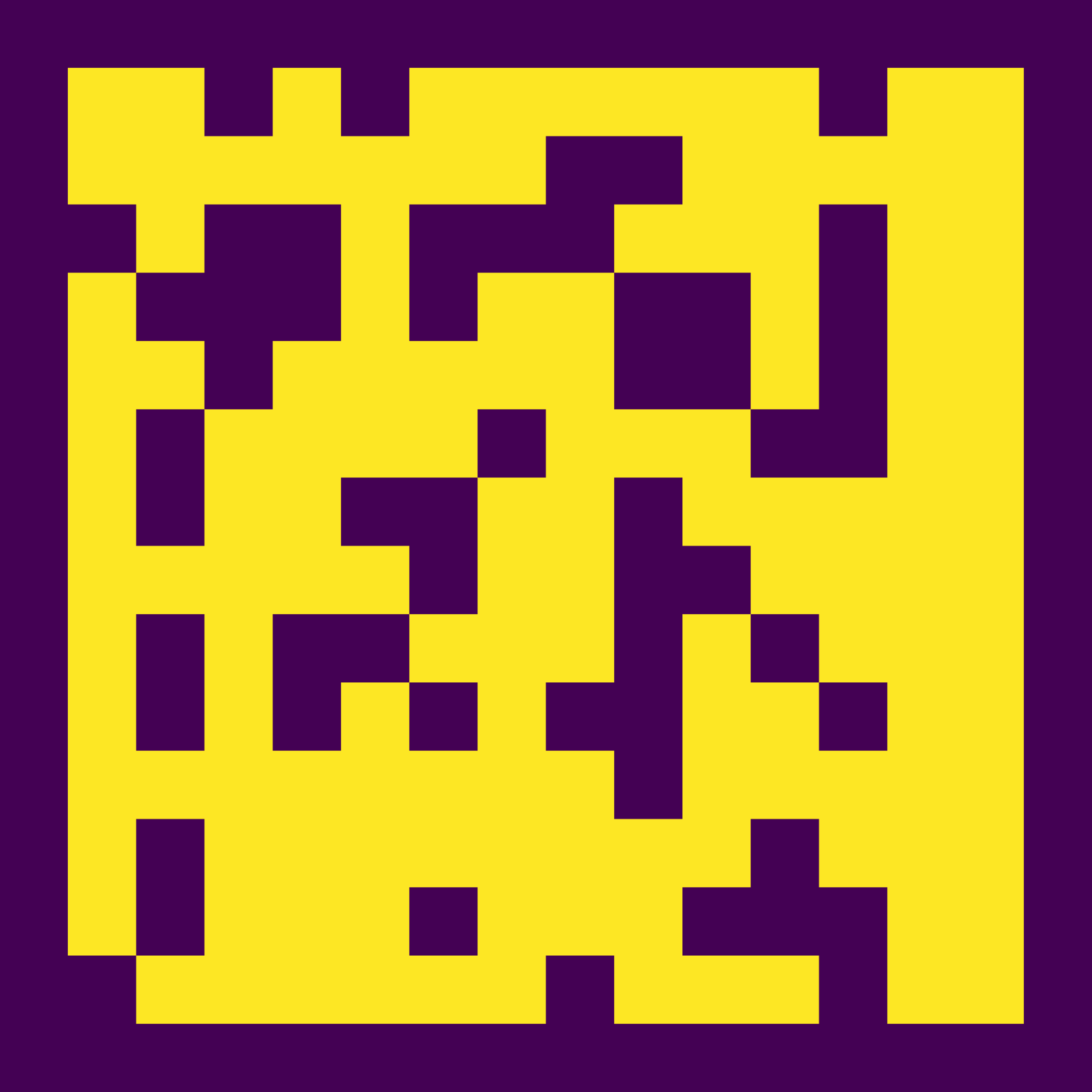}
         \includegraphics[width=0.24\textwidth]{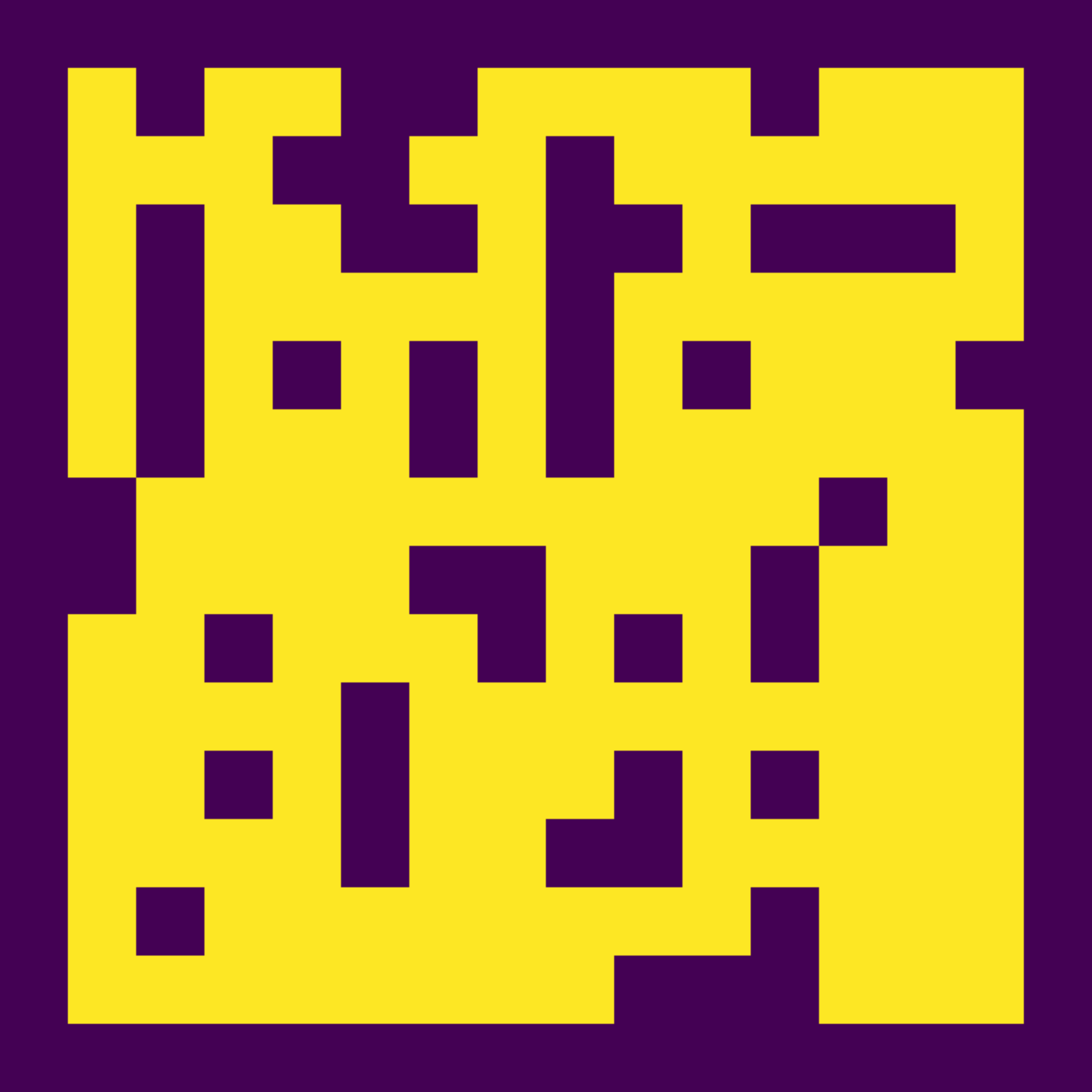}
         \includegraphics[width=0.24\textwidth]{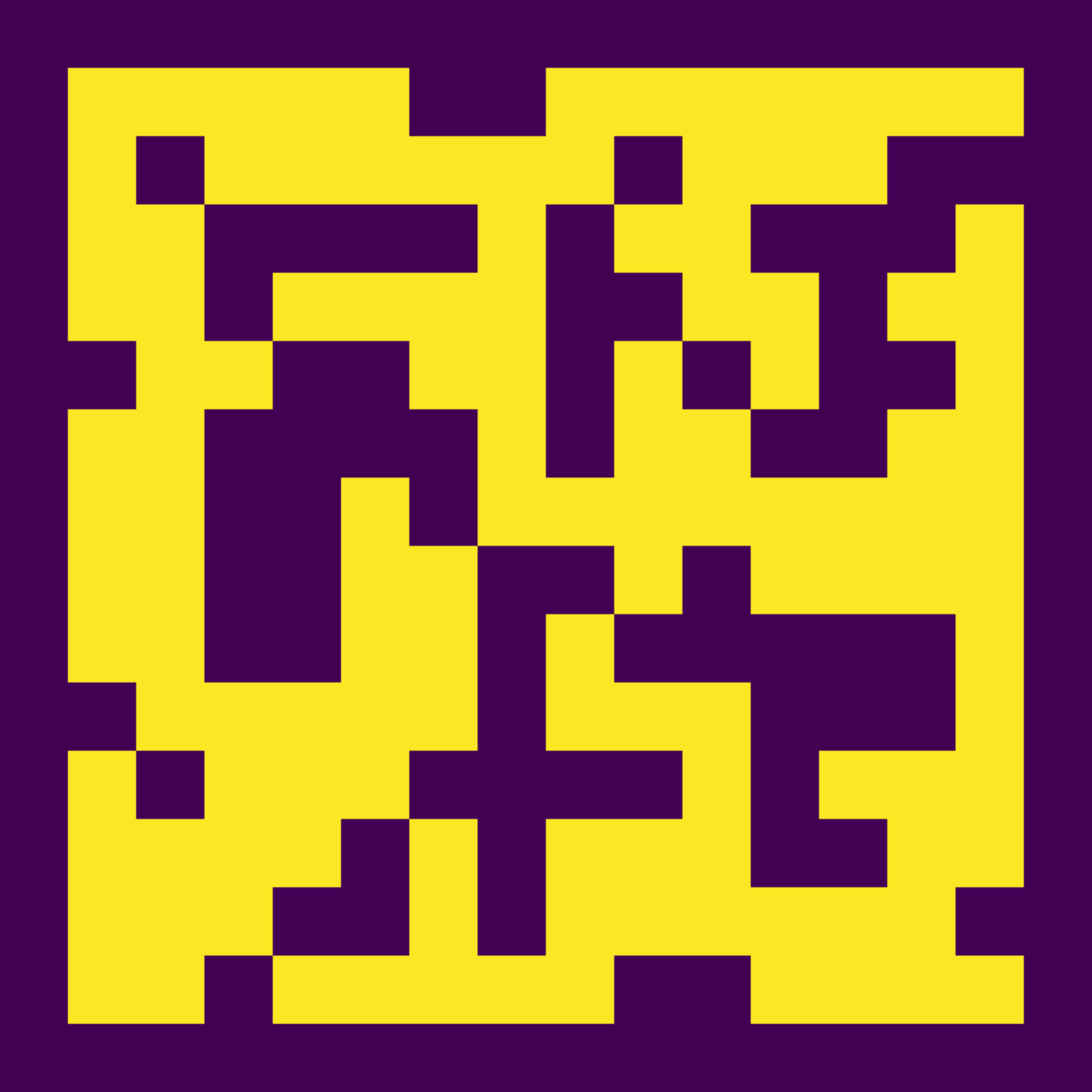}
         \includegraphics[width=0.24\textwidth]{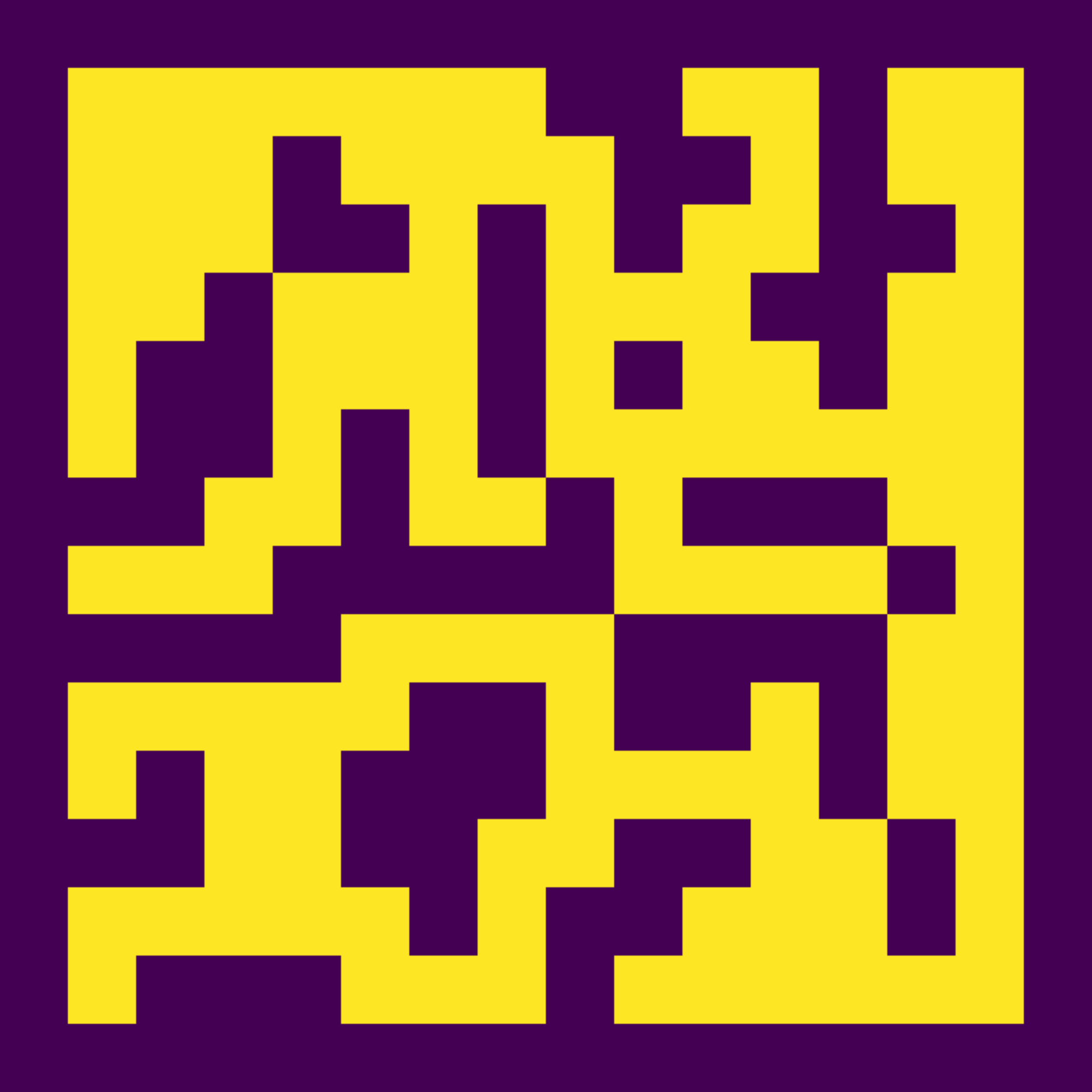}
         \caption{Normal-4}
         \label{fig:result_final4}
    \end{subfigure}
    \begin{subfigure}[t]{0.97\linewidth}
         \centering
         \includegraphics[width=0.24\textwidth]{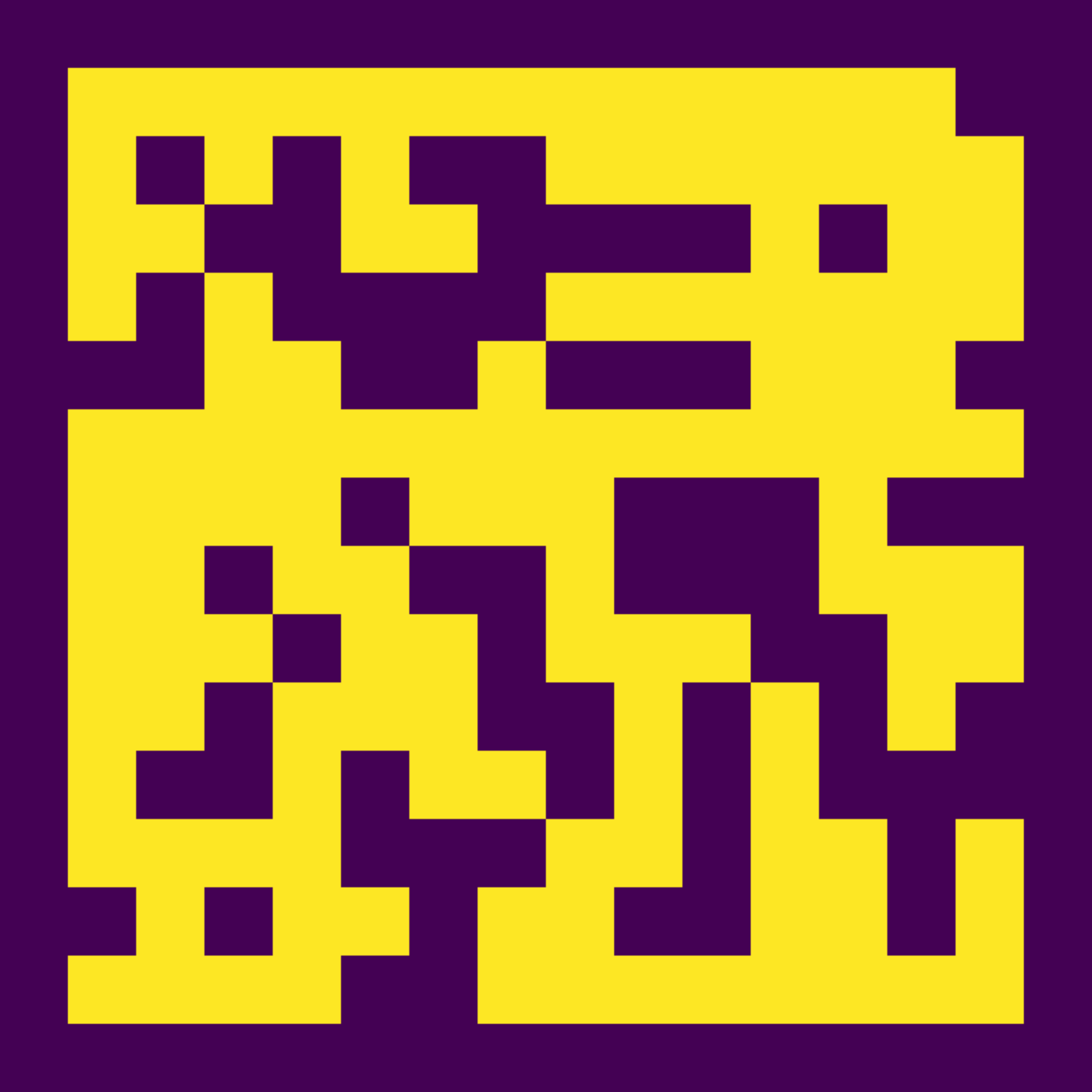}
         \includegraphics[width=0.24\textwidth]{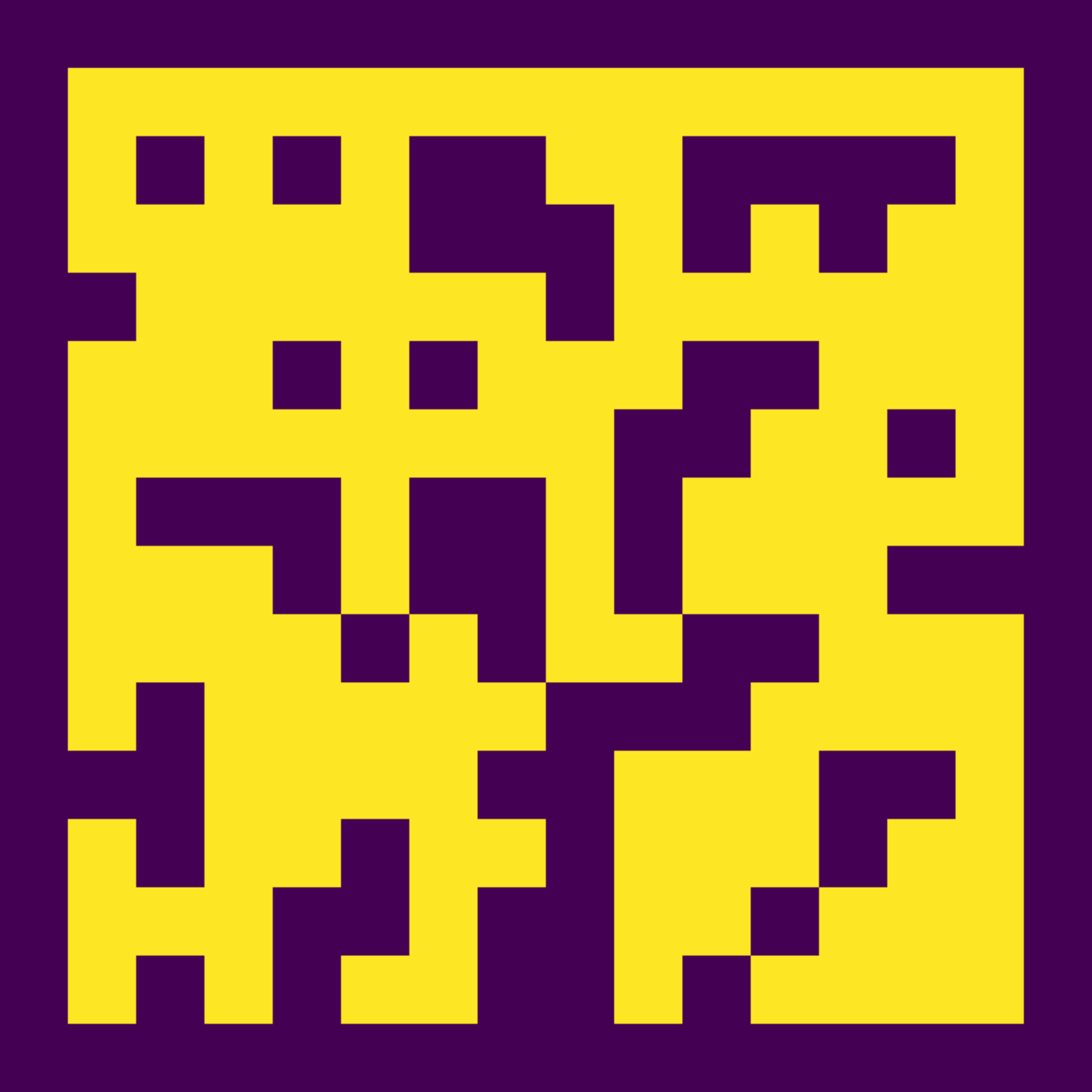}
         \includegraphics[width=0.24\textwidth]{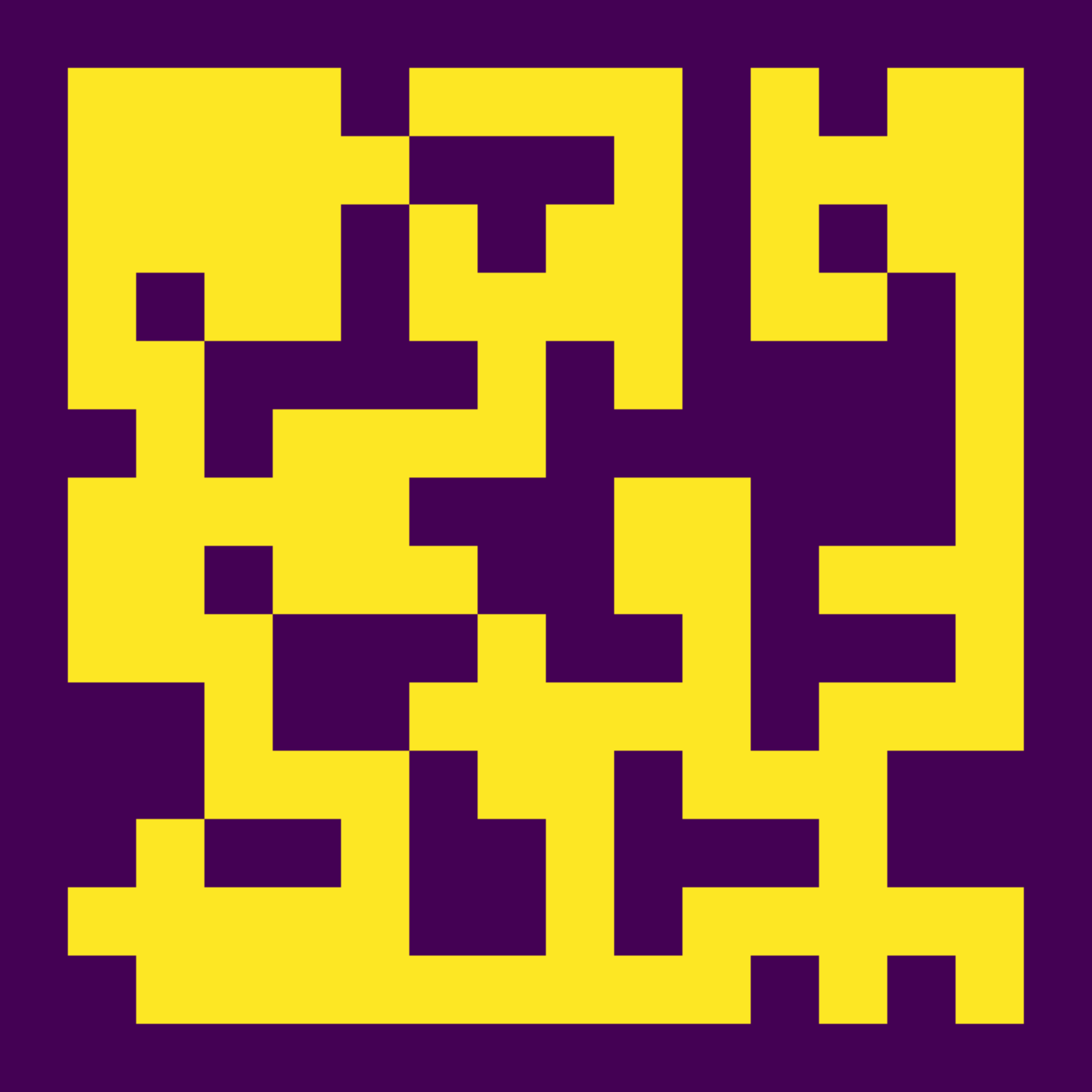}
         \includegraphics[width=0.24\textwidth]{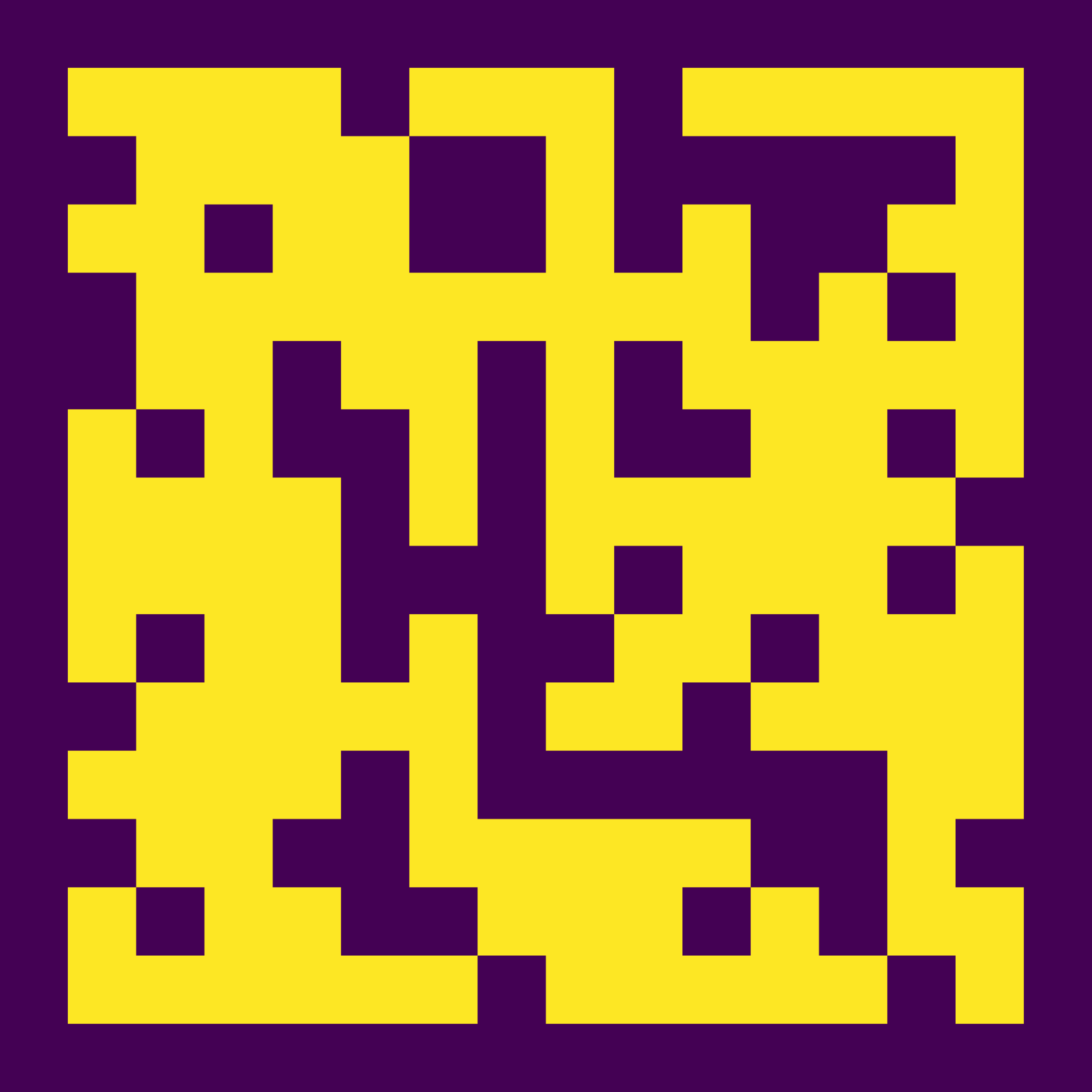}
         \caption{Normal-8}
         \label{fig:result_final8}
    \end{subfigure}
    \caption{Four successful maps generated from the different trained model via inference. None of these maps are generated using a fitness function. Each subcaption specifies the name of the experiment, divided into 2 parts: name and number. The name reflects the evolution process that created the dataset for this model to be trained on (``Normal'' or ``Assisted''). The number after the name of the experiment designates the number of epochs the network used for training.}
    \label{fig:generated_maps}
\end{figure}

\begin{figure*}
    \centering
    \begin{subfigure}[t]{0.23\textwidth}
         \centering
         \includegraphics[width=\textwidth]{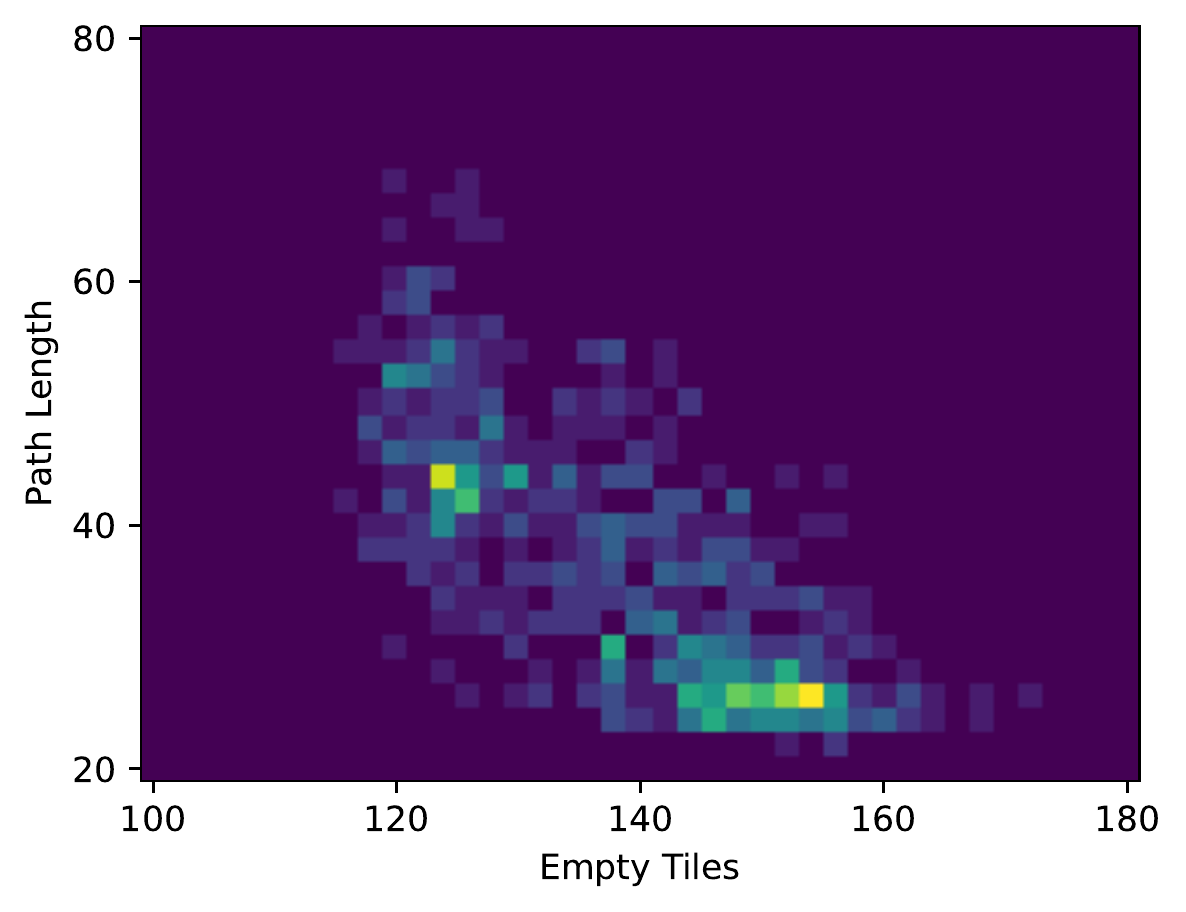}
         \caption{Assisted-2}
         \label{fig:div_assisted}
    \end{subfigure}
    \begin{subfigure}[t]{0.23\textwidth}
         \centering
         \includegraphics[width=\textwidth]{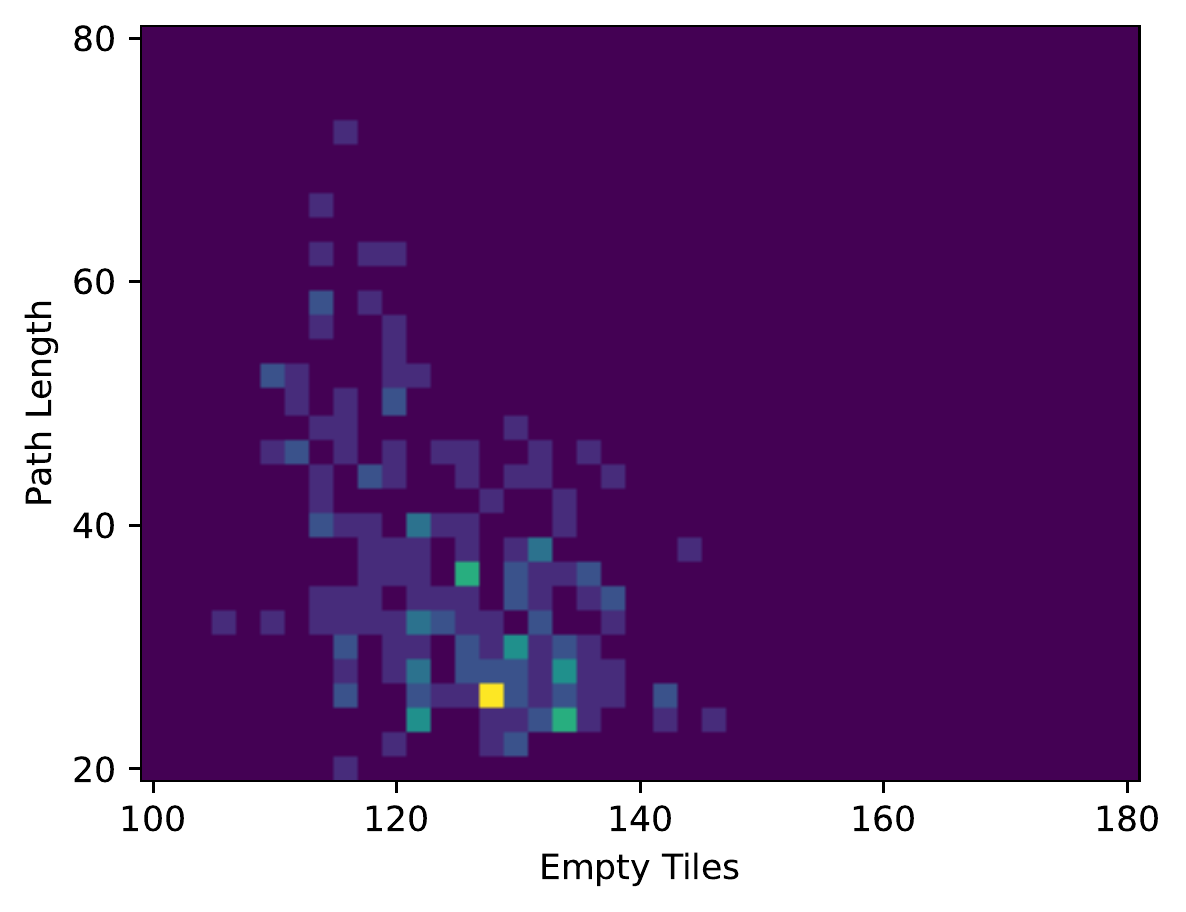}
         \caption{Normal-2}
         \label{fig:div_final2}
    \end{subfigure}
    \begin{subfigure}[t]{0.23\textwidth}
         \centering
         \includegraphics[width=\textwidth]{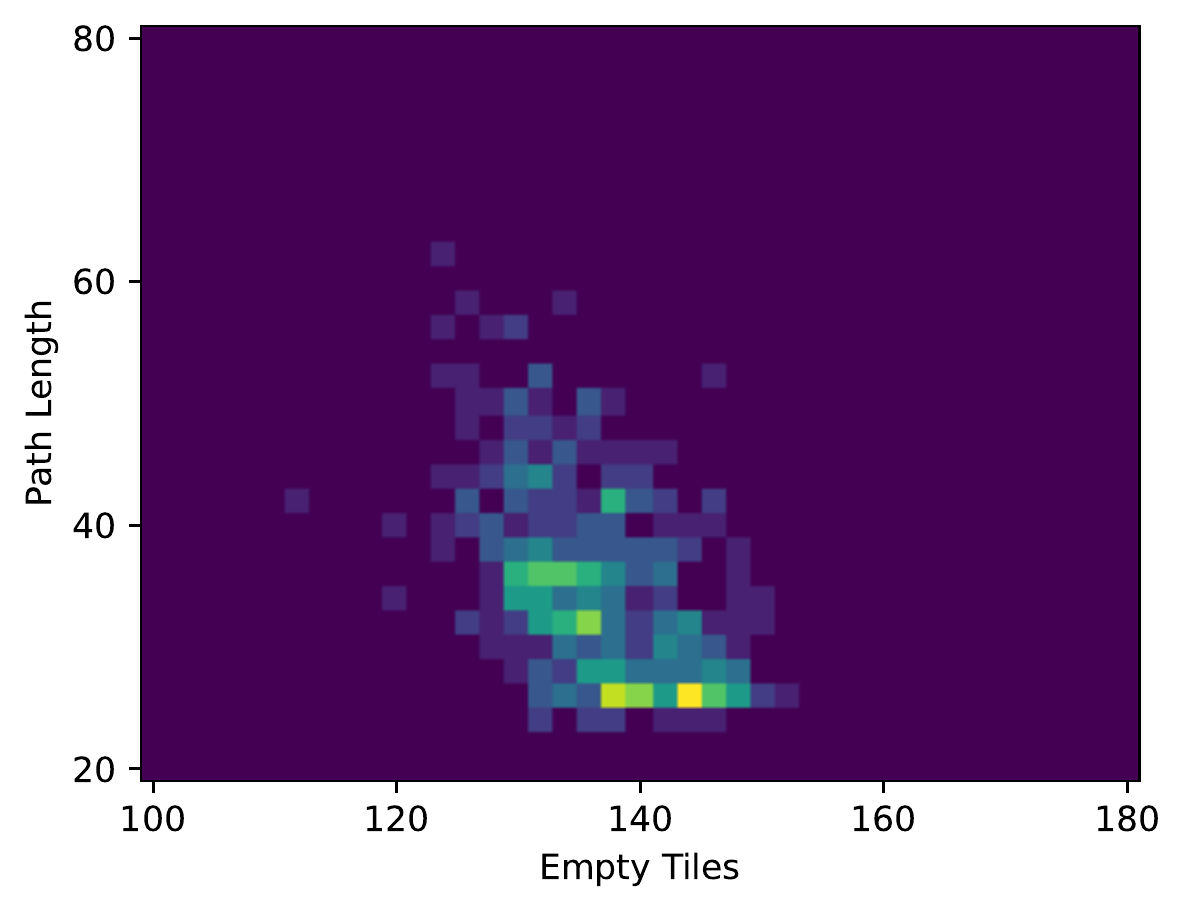}
         \caption{Normal-4}
         \label{fig:div_final4}
    \end{subfigure}
    \begin{subfigure}[t]{0.23\textwidth}
         \centering
         \includegraphics[width=\textwidth]{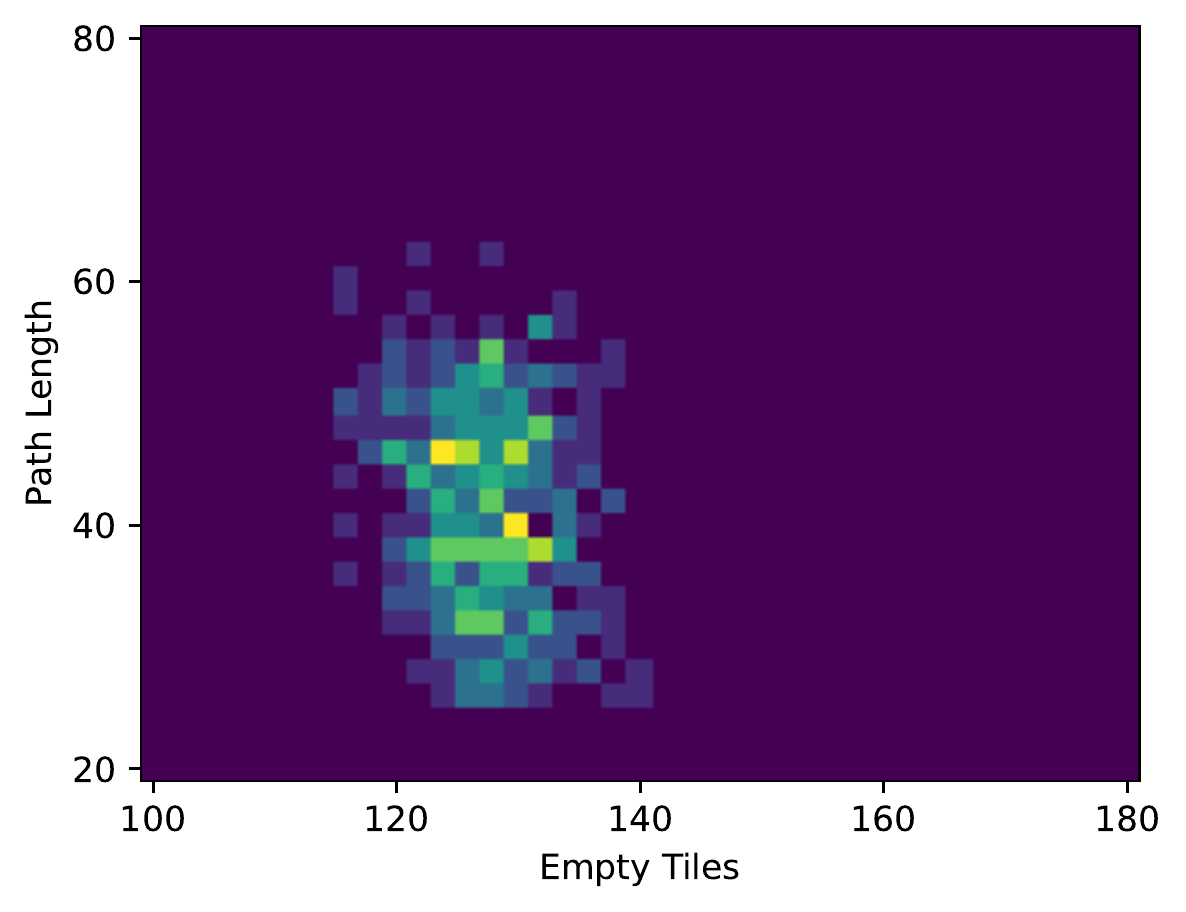}
         \caption{Normal-8}
         \label{fig:div_final8}
    \end{subfigure}
    \caption{Expressive range analysis of the generated maps during inference using the trained networks from the 4 different experiments. The expressive range plots all the successful generated maps on a 2D space, organized by the number of empty tiles and longest path length. Like Figure \ref{fig:generated_maps}, the name of experiment is divided into 2 parts: name and number. The name reflects the evolution process that created the dataset for this model to be trained on (``Normal'' or ``Assisted''). The `number after the name of the experiment designates the number of epochs the network used for training.}
    \label{fig:expressive_analysis}
\end{figure*}

All of the ``Normal'' trained models appear sensitive to the extracted dataset and varied widely in performance and diversity. We believe that using random distributions of mutations in the ``Normal'' method to build the evolution trajectories produces a noisy and unstable dataset to train on. However, trajectories built using the output of the trained network used by the ``Assisted'' method seem to produce a more consistent dataset that is easier to train a policy on. This is similar to on-policy reinforcement learning where the network is trying to learn from it own experience using rewards. In this work, the reward comes in the form of dataset filtering through evolution: only successful mutations are kept, and these mutations are the ones that the network will be trained on. It is not surprising that training for more epochs improves the performance of ``Normal'' method networks. However, this performance plateaus after only 4 epochs which confirms that the ``Assisted'' method helps training more than we expected. All of the networks take many iterations to converge on a successful level with minimum average of $18$ and maximum average of $61$. We believe that the ``Assisted-2'' may take less iterations if it is trained for more epochs with a result similar to what happens for the ``Normal'' methods.

Figure~\ref{fig:generated_maps} displays four successful maps from each experiment. The ``Assisted'' generated maps (Figure~\ref{fig:result_assisted}), have features similar to the original final maps. They contain long paths which contain long vertical lines. The ``Normal-2'' generated maps are more open in the upper left corner. We believe the small amount of epochs allows the network to learn a very simple policy which erases tiles to have full connectivity. ``Normal-4'' and ``Normal-8'' models also have more open space compared to ``Assisted'' but appeared more structured than ``Normal-2''. Often, the easiest way to reach full connectivity is to remove tiles from the level, as this will naturally ``open it up.'' We believe that ``Normal-2'' levels appear more open is because of the cascaded fitness function, where the evolution trajectory starts by connecting all the empty tiles then later improve path length. These evolution trajectories tend to have more ``Change to Empty'' actions at the beginning of evolution. We believe that the trained policy might have learned this behavior when the level is noisy. It learns to fully connect the map first, then later increase path length. This doesn't happen in the ``Assisted-2'' maps since the evolution is guided by a neural network that can take different actions which can be adjusted later. And since we terminate generation when everything is connected, ``Normal'' networks might achieve full connectivity just as generation ends, and before it begins to add solid tiles, giving the feeling of open space.

Finally, Figure~\ref{fig:expressive_analysis} displays the expressive range analysis of the trained models across 3 runs. Although ``Assisted-2'' generated maps have winding paths, they use on average more empty tiles compared to the rest of the models. Although it appears that the ``Normal'' method uses more empty tiles due to the openness of the maps, this is not always the case. For example, ``Normal-2'' maps seems more empty but a lot of the bottom left corner is solid. This is the same with ``Normal-4'' and ``Normal-8'' where most of the walls are clustered in the center. ``Normal'' maps give the feeling of openness while containing more solid tiles compared to the ``Assisted'' generated maps.

\section{Discussion}
We can contrast the mutation models method proposed in this paper to surrogate-assisted evolutionary computation. From a reinforcement learning perspective, fitness surrogate models can be thought of as models of the V-function, or in other words state value estimators. By using such a model to test multiple actions (mutations/crossover) the best action (surviving genome) can be chosen. The mutation models learned here are instead models of direct action selection. They are similar to models of the Q-function (action value estimators), but instead of returning the value of a particular mutation, they simply return the mutation to make (similar to actor networks).

Imitating mutation may not have the same ``speedup effect'' on evolution as surrogate modeling fitness functions as we still use the fitness function during evolution (which is computationally intensive) and the fitness value increases in the same rate as normal evolution (according to figure~\ref{fig:fitness}). However, they can provide expressive range where constructive generation methods fail to do so. Figure \ref{fig:expressive_analysis} shows that the ``Assisted-2'' method especially has a lot of promise in this area. Table \ref{tab:play_diverse_time} shows that the ``Assisted-2'' results in the highest map diversity. We use the top $10$ chromosomes of each generation to train with. Perhaps by expanding this, we might find even greater expressive range.

\begin{table}
    \centering
    \resizebox{\columnwidth}{!}{%
    \begin{tabular}{|l||c|c|c|}
    \hline
        Method & Success & Diversity & Wall Time (secs)\\
        \hline
        \hline
        Network & 99.67\% ± 0.49\% & 86.83\% ± 3.8\% & \textbf{0.6612 ± 2.3874} \\
        \hline
        Evolution & \textbf{100\%} & \textbf{96\%} &  12.6957 ± 2.2571\\
        \hline
    \end{tabular}%
    }
    \caption{The average and 95\% confidence interval of 100 generated levels by 3 networks trained using ``Assisted'' evolution for success rate, diversity, and wall clock time compared to the average for success rate, diversity, and wall clock time of running 100 normal evolutionary runs till the evolution find a fully connected level.}
    \label{tab:evolution_inference}
    \vspace{-5mm}
\end{table}

Search-based PCG methods are computationally expensive and are thus not commonly used for online generation. We ran an experiment to measure the wall clock time of normal evolution till it find a fully connected level or 2000 generations reached. The evolution took roughly 20x longer than mutation model inferencing as seen in table~\ref{tab:evolution_inference}. This difference would be even greater if we had used a more expensive fitness function during evolution, such as gameplay simulation. In other words, the speed advantage of Mutation Models will increase as the generation task gets harder. Learning generators negates the need for an evaluation function like fitness, thus speeding up the overall generation process during inference. Evaluation is what typically takes most of the computational time for search-based generators. Through the use of our method, developers might be able to benefit from the advantages that search-based PCG provides without having to bear the cost of online computation. Although evolution guarantees 100\% success rate and almost 100\% diversity as seen in table~\ref{tab:evolution_inference}, the ``Assisted'' trained networks have comparable success rate and diversity with drastically less time needed to generate the content.
    
We restricted the evolutionary algorithm to only have mutation to simplify the process of extracting trajectories. Removing crossover usually hinders evolution and in some problems the evolution might get stuck in local optima. This was not the case in our problem due to its simplicity. There is multiple different methods to extract trajectories while crossover is happening. These methods can be categorised into two main directions:
\begin{itemize}
    \item Converting the crossover into small mutation steps. When we extract trajectories whenever a crossover switch a big amount of tiles, we compare these tiles to the state before crossover and unpack these changes one a time in a random order (similar to path of destruction~\cite{siper2022path}).
    \item Consider mutation as a new starting point for a new trajectory. When we extract trajectories whenever a crossover happen we consider the current game level as new level and there is no dependency on the previous level.
\end{itemize}

This work introduces the opportunity to use an evolutionary algorithm as an assisted process for reinforcement learning, similar to Go-Explore algorithm~\cite{ecoffet2019go} and offline reinforcement learning~\cite{levine2020offline}. There is a lot to be explored and studied about this new paradigm. For example, what will happen if the evolution trajectories between the selected chromosomes are different? If we use a quality diversity algorithm such as MAP-Elites~\cite{mouret2015illuminating}, can we possibly learn a conditional mutator that can change the content towards a certain area in the generative space? We only discussed a very simple test bed problem, do the results from this test bed generalize to more complex games or even generic optimization problems? Also, what about the training algorithm, we only used basic supervised learning on trajectories but one could try using Backward algorithm similar to the one used in Go-Explore to get better results~\cite{ecoffet2019go}. Finally, we only explored creating the dataset from the evolution history, what if we can create the dataset using a similar technique to the Path of Destruction~\cite{siper2022path} where the evolution process is used as a method to generate the goal set. We believe that using something other than evolutionary history can free us to use more advanced evolutionary methods and/or indirect representation.

\section{Conclusion}
This paper proposes Mutation Models, a new method of generating content by building a machine learning model that imitates evolution. We use evolution trajectories to train a machine learning models in order to imitate mutation. This allows us to automatically create content generators by only specifying the fitness of the content, which is relatively domain-agnostic. The end result is an iterative content generator that is fast and does not need a fitness function during inference. Model building consists of two main loops (as shown in Figure~\ref{fig:system}): the evolution loop and training loop. The evolution loop is a normal evolutionary algorithm that generates content, while the training loop is responsible for extracting a training dataset from the top $X$ chromosomes. The training dataset is constructed by converting the evolution history into state-action pairs, on which a traditional supervised learning model is trained. We propose two methods for the framework: the ``Normal'' method (the machine learning model is trained after the evolution ends) and the ``Assisted'' method (the machine learning model is trained every $I$ generations and used to assist evolution). The results show that although the ``Assisted'' method does not help the evolution to be faster or more efficient, it does stabilize the trained networks, which learn better policies to imitate evolution in comparison to ``Normal'' method. Both methods can infer on randomly initialized maps and repair them without use of an evaluation function.


\bibliographystyle{ACM-Reference-Format}
\bibliography{biblo}


\begin{thebibliography}{46}


\ifx \showCODEN    \undefined \def \showCODEN     #1{\unskip}     \fi
\ifx \showDOI      \undefined \def \showDOI       #1{#1}\fi
\ifx \showISBNx    \undefined \def \showISBNx     #1{\unskip}     \fi
\ifx \showISBNxiii \undefined \def \showISBNxiii  #1{\unskip}     \fi
\ifx \showISSN     \undefined \def \showISSN      #1{\unskip}     \fi
\ifx \showLCCN     \undefined \def \showLCCN      #1{\unskip}     \fi
\ifx \shownote     \undefined \def \shownote      #1{#1}          \fi
\ifx \showarticletitle \undefined \def \showarticletitle #1{#1}   \fi
\ifx \showURL      \undefined \def \showURL       {\relax}        \fi
\providecommand\bibfield[2]{#2}
\providecommand\bibinfo[2]{#2}
\providecommand\natexlab[1]{#1}
\providecommand\showeprint[2][]{arXiv:#2}

\bibitem[Abboud and Schoenauer(2001)]%
        {abboud2001surrogate}
\bibfield{author}{\bibinfo{person}{Kamal Abboud} {and} \bibinfo{person}{Marc
  Schoenauer}.} \bibinfo{year}{2001}\natexlab{}.
\newblock \showarticletitle{Surrogate deterministic mutation: Preliminary
  results}. In \bibinfo{booktitle}{\emph{International Conference on Artificial
  Evolution}}. \bibinfo{publisher}{Springer}, \bibinfo{pages}{104--116}.
\newblock


\bibitem[Ashlock(2010)]%
        {ashlock2010elements}
\bibfield{author}{\bibinfo{person}{Daniel Ashlock}.}
  \bibinfo{year}{2010}\natexlab{}.
\newblock \showarticletitle{Automatic generation of game elements via
  evolution}. In \bibinfo{booktitle}{\emph{Conference on Computational
  Intelligence and Games}}. \bibinfo{publisher}{IEEE},
  \bibinfo{pages}{289--296}.
\newblock


\bibitem[Ashlock et~al\mbox{.}(2011)]%
        {ashlock2011search}
\bibfield{author}{\bibinfo{person}{Daniel Ashlock}, \bibinfo{person}{Colin
  Lee}, {and} \bibinfo{person}{Cameron McGuinness}.}
  \bibinfo{year}{2011}\natexlab{}.
\newblock \showarticletitle{Search-based procedural generation of maze-like
  levels}.
\newblock \bibinfo{journal}{\emph{Transactions on Computational Intelligence
  and AI in Games}} \bibinfo{volume}{3}, \bibinfo{number}{3}
  (\bibinfo{year}{2011}), \bibinfo{pages}{260--273}.
\newblock


\bibitem[Beyer and Schwefel(2002)]%
        {beyer2002evolution}
\bibfield{author}{\bibinfo{person}{Hans-Georg Beyer} {and}
  \bibinfo{person}{Hans-Paul Schwefel}.} \bibinfo{year}{2002}\natexlab{}.
\newblock \showarticletitle{Evolution strategies--a comprehensive
  introduction}.
\newblock \bibinfo{journal}{\emph{Natural computing}} \bibinfo{volume}{1},
  \bibinfo{number}{1} (\bibinfo{year}{2002}), \bibinfo{pages}{3--52}.
\newblock


\bibitem[Bhaumik et~al\mbox{.}(2020)]%
        {bhaumik2020tree}
\bibfield{author}{\bibinfo{person}{Debosmita Bhaumik}, \bibinfo{person}{Ahmed
  Khalifa}, \bibinfo{person}{Michael Green}, {and} \bibinfo{person}{Julian
  Togelius}.} \bibinfo{year}{2020}\natexlab{}.
\newblock \showarticletitle{Tree search versus optimization approaches for map
  generation}. In \bibinfo{booktitle}{\emph{Artificial Intelligence and
  Interactive Digital Entertainment}}, Vol.~\bibinfo{volume}{16}.
  \bibinfo{publisher}{AAAI}, \bibinfo{pages}{24--30}.
\newblock


\bibitem[Black(2021)]%
        {black2022ratcliff}
\bibfield{author}{\bibinfo{person}{Paul~E. Black}.}
  \bibinfo{year}{2021}\natexlab{}.
\newblock \bibinfo{booktitle}{\emph{Ratcliff/Obershelp pattern recognition}}.
\newblock
\urldef\tempurl%
\url{https://www.nist.gov/dads/HTML/ratcliffObershelp.html}
\showURL{%
\tempurl}


\bibitem[Browne and Maire(2010)]%
        {browne2010evolutionary}
\bibfield{author}{\bibinfo{person}{Cameron Browne} {and}
  \bibinfo{person}{Frederic Maire}.} \bibinfo{year}{2010}\natexlab{}.
\newblock \showarticletitle{Evolutionary game design}.
\newblock \bibinfo{journal}{\emph{Transactions on Computational Intelligence
  and AI in Games}} \bibinfo{volume}{2}, \bibinfo{number}{1}
  (\bibinfo{year}{2010}), \bibinfo{pages}{1--16}.
\newblock


\bibitem[Chen et~al\mbox{.}(2018)]%
        {chen2018q}
\bibfield{author}{\bibinfo{person}{Zhengxing Chen},
  \bibinfo{person}{Christopher Amato}, \bibinfo{person}{Truong-Huy~D Nguyen},
  \bibinfo{person}{Seth Cooper}, \bibinfo{person}{Yizhou Sun}, {and}
  \bibinfo{person}{Magy~Seif El-Nasr}.} \bibinfo{year}{2018}\natexlab{}.
\newblock \showarticletitle{Q-deckrec: A fast deck recommendation system for
  collectible card games}. In \bibinfo{booktitle}{\emph{Computational
  Intelligence and Games}}. \bibinfo{publisher}{IEEE}, \bibinfo{pages}{1--8}.
\newblock


\bibitem[Cook et~al\mbox{.}(2013)]%
        {cook2013mechanic}
\bibfield{author}{\bibinfo{person}{Michael Cook}, \bibinfo{person}{Simon
  Colton}, \bibinfo{person}{Azalea Raad}, {and} \bibinfo{person}{Jeremy Gow}.}
  \bibinfo{year}{2013}\natexlab{}.
\newblock \showarticletitle{Mechanic miner: Reflection-driven game mechanic
  discovery and level design}. In \bibinfo{booktitle}{\emph{European Conference
  on the Applications of Evolutionary Computation}}.
  \bibinfo{publisher}{Springer}, \bibinfo{pages}{284--293}.
\newblock


\bibitem[Dahlskog et~al\mbox{.}(2014)]%
        {dahlskog2014linear}
\bibfield{author}{\bibinfo{person}{Steve Dahlskog}, \bibinfo{person}{Julian
  Togelius}, {and} \bibinfo{person}{Mark~J Nelson}.}
  \bibinfo{year}{2014}\natexlab{}.
\newblock \showarticletitle{Linear levels through n-grams}. In
  \bibinfo{booktitle}{\emph{International Academic MindTrek Conference: Media
  Business, Management, Content \& Services}}. \bibinfo{publisher}{ACM},
  \bibinfo{pages}{200--206}.
\newblock


\bibitem[Delarosa et~al\mbox{.}(2021)]%
        {delarosa2021mixed}
\bibfield{author}{\bibinfo{person}{Omar Delarosa}, \bibinfo{person}{Hang Dong},
  \bibinfo{person}{Mindy Ruan}, \bibinfo{person}{Ahmed Khalifa}, {and}
  \bibinfo{person}{Julian Togelius}.} \bibinfo{year}{2021}\natexlab{}.
\newblock \showarticletitle{Mixed-initiative level design with rl brush}. In
  \bibinfo{booktitle}{\emph{Conference on Computational Intelligence in Music,
  Sound, Art and Design}}. \bibinfo{publisher}{Springer},
  \bibinfo{pages}{412--426}.
\newblock


\bibitem[Earle et~al\mbox{.}(2021)]%
        {earle2021learning}
\bibfield{author}{\bibinfo{person}{Sam Earle}, \bibinfo{person}{Maria Edwards},
  \bibinfo{person}{Ahmed Khalifa}, \bibinfo{person}{Philip Bontrager}, {and}
  \bibinfo{person}{Julian Togelius}.} \bibinfo{year}{2021}\natexlab{}.
\newblock \showarticletitle{Learning controllable content generators}. In
  \bibinfo{booktitle}{\emph{Conference on Games}}. \bibinfo{publisher}{IEEE},
  \bibinfo{pages}{1--9}.
\newblock


\bibitem[Earle et~al\mbox{.}(2022)]%
        {earle2022illuminating}
\bibfield{author}{\bibinfo{person}{Sam Earle}, \bibinfo{person}{Justin Snider},
  \bibinfo{person}{Matthew~C Fontaine}, \bibinfo{person}{Stefanos Nikolaidis},
  {and} \bibinfo{person}{Julian Togelius}.} \bibinfo{year}{2022}\natexlab{}.
\newblock \showarticletitle{Illuminating diverse neural cellular automata for
  level generation}. In \bibinfo{booktitle}{\emph{Genetic and Evolutionary
  Computation Conference}}. \bibinfo{publisher}{ACM}, \bibinfo{pages}{68--76}.
\newblock


\bibitem[Ecoffet et~al\mbox{.}(2019)]%
        {ecoffet2019go}
\bibfield{author}{\bibinfo{person}{Adrien Ecoffet}, \bibinfo{person}{Joost
  Huizinga}, \bibinfo{person}{Joel Lehman}, \bibinfo{person}{Kenneth~O
  Stanley}, {and} \bibinfo{person}{Jeff Clune}.}
  \bibinfo{year}{2019}\natexlab{}.
\newblock \showarticletitle{Go-explore: a new approach for hard-exploration
  problems}.
\newblock \bibinfo{journal}{\emph{arXiv preprint arXiv:1901.10995}}
  (\bibinfo{year}{2019}).
\newblock


\bibitem[Grinblat(2016)]%
        {griblat2016caves}
\bibfield{author}{\bibinfo{person}{Jason Grinblat}.}
  \bibinfo{year}{2016}\natexlab{}.
\newblock \bibinfo{booktitle}{\emph{Markov by candlelight}}.
\newblock
\urldef\tempurl%
\url{https://www.youtube.com/watch?v=3AjlsTtrfVY}
\showURL{%
\tempurl}


\bibitem[Guzdial et~al\mbox{.}(2019)]%
        {guzdial2019friend}
\bibfield{author}{\bibinfo{person}{Matthew Guzdial}, \bibinfo{person}{Nicholas
  Liao}, \bibinfo{person}{Jonathan Chen}, \bibinfo{person}{Shao-Yu Chen},
  \bibinfo{person}{Shukan Shah}, \bibinfo{person}{Vishwa Shah},
  \bibinfo{person}{Joshua Reno}, \bibinfo{person}{Gillian Smith}, {and}
  \bibinfo{person}{Mark~O Riedl}.} \bibinfo{year}{2019}\natexlab{}.
\newblock \showarticletitle{Friend, collaborator, student, manager: How design
  of an ai-driven game level editor affects creators}. In
  \bibinfo{booktitle}{\emph{CHI conference on human factors in computing
  systems}}. \bibinfo{publisher}{ACM}, \bibinfo{pages}{1--13}.
\newblock


\bibitem[Guzdial et~al\mbox{.}(2018)]%
        {guzdial2018co}
\bibfield{author}{\bibinfo{person}{Matthew Guzdial}, \bibinfo{person}{Nicholas
  Liao}, {and} \bibinfo{person}{Mark Riedl}.} \bibinfo{year}{2018}\natexlab{}.
\newblock \showarticletitle{Co-creative level design via machine learning}. In
  \bibinfo{booktitle}{\emph{AIIDE workshop on Experimental AI in Games}}.
  \bibinfo{publisher}{AAAI}.
\newblock


\bibitem[Jain et~al\mbox{.}(2016)]%
        {jain2016autoencoders}
\bibfield{author}{\bibinfo{person}{Rishabh Jain}, \bibinfo{person}{Aaron
  Isaksen}, \bibinfo{person}{Christoffer Holmg{\aa}rd}, {and}
  \bibinfo{person}{Julian Togelius}.} \bibinfo{year}{2016}\natexlab{}.
\newblock \showarticletitle{Autoencoders for level generation, repair, and
  recognition}. In \bibinfo{booktitle}{\emph{ICCC workshop on computational
  creativity and games}}, Vol.~\bibinfo{volume}{9}. \bibinfo{publisher}{IEEE}.
\newblock


\bibitem[Jin(2011)]%
        {jin2011surrogate}
\bibfield{author}{\bibinfo{person}{Yaochu Jin}.}
  \bibinfo{year}{2011}\natexlab{}.
\newblock \showarticletitle{Surrogate-assisted evolutionary computation: Recent
  advances and future challenges}.
\newblock \bibinfo{journal}{\emph{Swarm and Evolutionary Computation}}
  \bibinfo{volume}{1}, \bibinfo{number}{2} (\bibinfo{year}{2011}),
  \bibinfo{pages}{61--70}.
\newblock


\bibitem[Karth and Smith(2017)]%
        {karth2017wavefunctioncollapse}
\bibfield{author}{\bibinfo{person}{Isaac Karth} {and} \bibinfo{person}{Adam~M
  Smith}.} \bibinfo{year}{2017}\natexlab{}.
\newblock \showarticletitle{WaveFunctionCollapse is constraint solving in the
  wild}. In \bibinfo{booktitle}{\emph{Foundations of Digital Games}}.
  \bibinfo{publisher}{ACM}, \bibinfo{pages}{1--10}.
\newblock


\bibitem[Kerssemakers et~al\mbox{.}(2012)]%
        {kerssemakers2012procedural}
\bibfield{author}{\bibinfo{person}{Manuel Kerssemakers}, \bibinfo{person}{Jeppe
  Tuxen}, \bibinfo{person}{Julian Togelius}, {and} \bibinfo{person}{Georgios~N
  Yannakakis}.} \bibinfo{year}{2012}\natexlab{}.
\newblock \showarticletitle{A procedural procedural level generator generator}.
  In \bibinfo{booktitle}{\emph{Computational Intelligence and Games}}.
  \bibinfo{publisher}{IEEE}, \bibinfo{pages}{335--341}.
\newblock


\bibitem[Khalifa et~al\mbox{.}(2020)]%
        {khalifa2020pcgrl}
\bibfield{author}{\bibinfo{person}{Ahmed Khalifa}, \bibinfo{person}{Philip
  Bontrager}, \bibinfo{person}{Sam Earle}, {and} \bibinfo{person}{Julian
  Togelius}.} \bibinfo{year}{2020}\natexlab{}.
\newblock \showarticletitle{Pcgrl: Procedural content generation via
  reinforcement learning}. In \bibinfo{booktitle}{\emph{Artificial Intelligence
  and Interactive Digital Entertainment}}, Vol.~\bibinfo{volume}{16}.
  \bibinfo{publisher}{AAAI}, \bibinfo{pages}{95--101}.
\newblock


\bibitem[Khalifa and Fayek(2015)]%
        {khalifa2015automatic}
\bibfield{author}{\bibinfo{person}{Ahmed Khalifa} {and} \bibinfo{person}{Magda
  Fayek}.} \bibinfo{year}{2015}\natexlab{}.
\newblock \showarticletitle{Automatic puzzle level generation: A general
  approach using a description language}. In \bibinfo{booktitle}{\emph{ICCC
  workshop on computational creativity and games}}. \bibinfo{publisher}{IEEE}.
\newblock


\bibitem[Khalifa et~al\mbox{.}(2017)]%
        {khalifa2017general}
\bibfield{author}{\bibinfo{person}{Ahmed Khalifa},
  \bibinfo{person}{Michael~Cerny Green}, \bibinfo{person}{Diego Perez-Liebana},
  {and} \bibinfo{person}{Julian Togelius}.} \bibinfo{year}{2017}\natexlab{}.
\newblock \showarticletitle{General video game rule generation}. In
  \bibinfo{booktitle}{\emph{Computational Intelligence and Games}}.
  \bibinfo{publisher}{IEEE}, \bibinfo{pages}{170--177}.
\newblock


\bibitem[Khalifa and Togelius(2020)]%
        {khalifa2020multi}
\bibfield{author}{\bibinfo{person}{Ahmed Khalifa} {and} \bibinfo{person}{Julian
  Togelius}.} \bibinfo{year}{2020}\natexlab{}.
\newblock \showarticletitle{Multi-Objective level generator generation with
  Marahel}. In \bibinfo{booktitle}{\emph{Foundations of Digital Games}}.
  \bibinfo{publisher}{ACM}, \bibinfo{pages}{1--8}.
\newblock


\bibitem[Levine et~al\mbox{.}(2020)]%
        {levine2020offline}
\bibfield{author}{\bibinfo{person}{Sergey Levine}, \bibinfo{person}{Aviral
  Kumar}, \bibinfo{person}{George Tucker}, {and} \bibinfo{person}{Justin Fu}.}
  \bibinfo{year}{2020}\natexlab{}.
\newblock \showarticletitle{Offline reinforcement learning: Tutorial, review,
  and perspectives on open problems}.
\newblock \bibinfo{journal}{\emph{arXiv preprint arXiv:2005.01643}}
  (\bibinfo{year}{2020}).
\newblock


\bibitem[Liu et~al\mbox{.}(2021)]%
        {liu2021deep}
\bibfield{author}{\bibinfo{person}{Jialin Liu}, \bibinfo{person}{Sam
  Snodgrass}, \bibinfo{person}{Ahmed Khalifa}, \bibinfo{person}{Sebastian
  Risi}, \bibinfo{person}{Georgios~N Yannakakis}, {and} \bibinfo{person}{Julian
  Togelius}.} \bibinfo{year}{2021}\natexlab{}.
\newblock \showarticletitle{Deep learning for procedural content generation}.
\newblock \bibinfo{journal}{\emph{Neural Computing and Applications}}
  \bibinfo{volume}{33}, \bibinfo{number}{1} (\bibinfo{year}{2021}),
  \bibinfo{pages}{19--37}.
\newblock


\bibitem[Mahmoudi-Nejad et~al\mbox{.}(2021)]%
        {mahmoudi2021arachnophobia}
\bibfield{author}{\bibinfo{person}{Athar Mahmoudi-Nejad},
  \bibinfo{person}{Matthew Guzdial}, {and} \bibinfo{person}{Pierre Boulanger}.}
  \bibinfo{year}{2021}\natexlab{}.
\newblock \showarticletitle{Arachnophobia exposure therapy using
  experience-driven procedural content generation via reinforcement learning
  (EDPCGRL)}. In \bibinfo{booktitle}{\emph{Artificial Intelligence and
  Interactive Digital Entertainment}}, Vol.~\bibinfo{volume}{17}.
  \bibinfo{publisher}{AAAI}, \bibinfo{pages}{164--171}.
\newblock


\bibitem[Mnih et~al\mbox{.}(2013)]%
        {mnih2013playing}
\bibfield{author}{\bibinfo{person}{Volodymyr Mnih}, \bibinfo{person}{Koray
  Kavukcuoglu}, \bibinfo{person}{David Silver}, \bibinfo{person}{Alex Graves},
  \bibinfo{person}{Ioannis Antonoglou}, \bibinfo{person}{Daan Wierstra}, {and}
  \bibinfo{person}{Martin Riedmiller}.} \bibinfo{year}{2013}\natexlab{}.
\newblock \showarticletitle{Playing atari with deep reinforcement learning}.
\newblock \bibinfo{journal}{\emph{arXiv preprint arXiv:1312.5602}}
  (\bibinfo{year}{2013}).
\newblock


\bibitem[Mouret and Clune(2015)]%
        {mouret2015illuminating}
\bibfield{author}{\bibinfo{person}{Jean-Baptiste Mouret} {and}
  \bibinfo{person}{Jeff Clune}.} \bibinfo{year}{2015}\natexlab{}.
\newblock \showarticletitle{Illuminating search spaces by mapping elites}.
\newblock \bibinfo{journal}{\emph{arXiv preprint arXiv:1504.04909}}
  (\bibinfo{year}{2015}).
\newblock


\bibitem[Nam and Ikeda(2019)]%
        {nam2019generation}
\bibfield{author}{\bibinfo{person}{SangGyu Nam} {and} \bibinfo{person}{Kokolo
  Ikeda}.} \bibinfo{year}{2019}\natexlab{}.
\newblock \showarticletitle{Generation of diverse stages in turn-based
  role-playing game using reinforcement learning}. In
  \bibinfo{booktitle}{\emph{Conference on Games}}. IEEE, \bibinfo{pages}{1--8}.
\newblock


\bibitem[Ong et~al\mbox{.}(2003)]%
        {ong2003evolutionary}
\bibfield{author}{\bibinfo{person}{Yew~S Ong}, \bibinfo{person}{Prasanth~B
  Nair}, {and} \bibinfo{person}{Andrew~J Keane}.}
  \bibinfo{year}{2003}\natexlab{}.
\newblock \showarticletitle{Evolutionary optimization of computationally
  expensive problems via surrogate modeling}.
\newblock \bibinfo{journal}{\emph{AIAA journal}} \bibinfo{volume}{41},
  \bibinfo{number}{4} (\bibinfo{year}{2003}), \bibinfo{pages}{687--696}.
\newblock


\bibitem[Perez-Liebana et~al\mbox{.}(2019)]%
        {perez2019general}
\bibfield{author}{\bibinfo{person}{Diego Perez-Liebana},
  \bibinfo{person}{Jialin Liu}, \bibinfo{person}{Ahmed Khalifa},
  \bibinfo{person}{Raluca~D Gaina}, \bibinfo{person}{Julian Togelius}, {and}
  \bibinfo{person}{Simon~M Lucas}.} \bibinfo{year}{2019}\natexlab{}.
\newblock \showarticletitle{General video game ai: A multitrack framework for
  evaluating agents, games, and content generation algorithms}.
\newblock \bibinfo{journal}{\emph{Transactions on Games}} \bibinfo{volume}{11},
  \bibinfo{number}{3} (\bibinfo{year}{2019}), \bibinfo{pages}{195--214}.
\newblock


\bibitem[Sarkar and Cooper(2021)]%
        {sarkar2021dungeon}
\bibfield{author}{\bibinfo{person}{Anurag Sarkar} {and} \bibinfo{person}{Seth
  Cooper}.} \bibinfo{year}{2021}\natexlab{}.
\newblock \showarticletitle{Dungeon and Platformer Level Blending and
  Generation using Conditional VAEs}. In \bibinfo{booktitle}{\emph{Conference
  on Games}}. \bibinfo{publisher}{IEEE}, \bibinfo{pages}{1--8}.
\newblock


\bibitem[Shaker et~al\mbox{.}(2013)]%
        {shaker2013evolving}
\bibfield{author}{\bibinfo{person}{Noor Shaker}, \bibinfo{person}{Mohammad
  Shaker}, {and} \bibinfo{person}{Julian Togelius}.}
  \bibinfo{year}{2013}\natexlab{}.
\newblock \showarticletitle{Evolving playable content for cut the rope through
  a simulation-based approach}. In \bibinfo{booktitle}{\emph{Artificial
  Intelligence and Interactive Digital Entertainment Conference}}.
  \bibinfo{publisher}{AAAI}.
\newblock


\bibitem[Shu et~al\mbox{.}(2021)]%
        {shu2021experience}
\bibfield{author}{\bibinfo{person}{Tianye Shu}, \bibinfo{person}{Jialin Liu},
  {and} \bibinfo{person}{Georgios~N Yannakakis}.}
  \bibinfo{year}{2021}\natexlab{}.
\newblock \showarticletitle{Experience-driven PCG via reinforcement learning: A
  Super Mario Bros study}. In \bibinfo{booktitle}{\emph{Conference on Games}}.
  \bibinfo{publisher}{IEEE}, \bibinfo{pages}{1--9}.
\newblock


\bibitem[Siper et~al\mbox{.}(2022)]%
        {siper2022path}
\bibfield{author}{\bibinfo{person}{Matthew Siper}, \bibinfo{person}{Ahmed
  Khalifa}, {and} \bibinfo{person}{Julian Togelius}.}
  \bibinfo{year}{2022}\natexlab{}.
\newblock \showarticletitle{Path of Destruction: Learning an Iterative Level
  Generator Using a Small Dataset}.
\newblock \bibinfo{journal}{\emph{arXiv preprint arXiv:2202.10184}}
  (\bibinfo{year}{2022}).
\newblock


\bibitem[Snodgrass and Ontan{\'o}n(2016)]%
        {snodgrass2016learning}
\bibfield{author}{\bibinfo{person}{Sam Snodgrass} {and}
  \bibinfo{person}{Santiago Ontan{\'o}n}.} \bibinfo{year}{2016}\natexlab{}.
\newblock \showarticletitle{Learning to generate video game maps using markov
  models}.
\newblock \bibinfo{journal}{\emph{Transactions on computational intelligence
  and AI in games}} \bibinfo{volume}{9}, \bibinfo{number}{4}
  (\bibinfo{year}{2016}), \bibinfo{pages}{410--422}.
\newblock


\bibitem[Summerville et~al\mbox{.}(2016)]%
        {summerville2016learning}
\bibfield{author}{\bibinfo{person}{Adam Summerville}, \bibinfo{person}{Matthew
  Guzdial}, \bibinfo{person}{Michael Mateas}, {and} \bibinfo{person}{Mark~O
  Riedl}.} \bibinfo{year}{2016}\natexlab{}.
\newblock \showarticletitle{Learning player tailored content from observation:
  Platformer level generation from video traces using lstms}. In
  \bibinfo{booktitle}{\emph{Artificial intelligence and interactive digital
  entertainment conference}}. \bibinfo{publisher}{AAAI}.
\newblock


\bibitem[Summerville et~al\mbox{.}(2018)]%
        {summerville2018procedural}
\bibfield{author}{\bibinfo{person}{Adam Summerville}, \bibinfo{person}{Sam
  Snodgrass}, \bibinfo{person}{Matthew Guzdial}, \bibinfo{person}{Christoffer
  Holmg{\aa}rd}, \bibinfo{person}{Amy~K Hoover}, \bibinfo{person}{Aaron
  Isaksen}, \bibinfo{person}{Andy Nealen}, {and} \bibinfo{person}{Julian
  Togelius}.} \bibinfo{year}{2018}\natexlab{}.
\newblock \showarticletitle{Procedural content generation via machine learning
  (PCGML)}.
\newblock \bibinfo{journal}{\emph{Transactions on Games}} \bibinfo{volume}{10},
  \bibinfo{number}{3} (\bibinfo{year}{2018}), \bibinfo{pages}{257--270}.
\newblock


\bibitem[Togelius et~al\mbox{.}(2011)]%
        {togelius2011search}
\bibfield{author}{\bibinfo{person}{Julian Togelius},
  \bibinfo{person}{Georgios~N Yannakakis}, \bibinfo{person}{Kenneth~O Stanley},
  {and} \bibinfo{person}{Cameron Browne}.} \bibinfo{year}{2011}\natexlab{}.
\newblock \showarticletitle{Search-based procedural content generation: A
  taxonomy and survey}.
\newblock \bibinfo{journal}{\emph{Transactions on Computational Intelligence
  and AI in Games}} \bibinfo{volume}{3}, \bibinfo{number}{3}
  (\bibinfo{year}{2011}), \bibinfo{pages}{172--186}.
\newblock


\bibitem[Torrado et~al\mbox{.}(2020)]%
        {torrado2020bootstrapping}
\bibfield{author}{\bibinfo{person}{Ruben~Rodriguez Torrado},
  \bibinfo{person}{Ahmed Khalifa}, \bibinfo{person}{Michael~Cerny Green},
  \bibinfo{person}{Niels Justesen}, \bibinfo{person}{Sebastian Risi}, {and}
  \bibinfo{person}{Julian Togelius}.} \bibinfo{year}{2020}\natexlab{}.
\newblock \showarticletitle{Bootstrapping conditional gans for video game level
  generation}. In \bibinfo{booktitle}{\emph{Conference on Games}}.
  \bibinfo{publisher}{IEEE}, \bibinfo{pages}{41--48}.
\newblock


\bibitem[Volz et~al\mbox{.}(2018)]%
        {volz2018evolving}
\bibfield{author}{\bibinfo{person}{Vanessa Volz}, \bibinfo{person}{Jacob
  Schrum}, \bibinfo{person}{Jialin Liu}, \bibinfo{person}{Simon~M Lucas},
  \bibinfo{person}{Adam Smith}, {and} \bibinfo{person}{Sebastian Risi}.}
  \bibinfo{year}{2018}\natexlab{}.
\newblock \showarticletitle{Evolving mario levels in the latent space of a deep
  convolutional generative adversarial network}. In
  \bibinfo{booktitle}{\emph{Genetic and evolutionary computation conference}}.
  \bibinfo{publisher}{ACM}, \bibinfo{pages}{221--228}.
\newblock


\bibitem[Werneck and Clua(2020)]%
        {werneck2020generating}
\bibfield{author}{\bibinfo{person}{Mariana Werneck} {and}
  \bibinfo{person}{Esteban~WG Clua}.} \bibinfo{year}{2020}\natexlab{}.
\newblock \showarticletitle{Generating procedural dungeons using machine
  learning methods}. In \bibinfo{booktitle}{\emph{Brazilian Symposium on
  Computer Games and Digital Entertainment}}. IEEE, \bibinfo{pages}{90--96}.
\newblock


\bibitem[Ye et~al\mbox{.}(2020)]%
        {ye2020rotation}
\bibfield{author}{\bibinfo{person}{Chang Ye}, \bibinfo{person}{Ahmed Khalifa},
  \bibinfo{person}{Philip Bontrager}, {and} \bibinfo{person}{Julian Togelius}.}
  \bibinfo{year}{2020}\natexlab{}.
\newblock \showarticletitle{Rotation, translation, and cropping for zero-shot
  generalization}. In \bibinfo{booktitle}{\emph{Conference on Games}}.
  \bibinfo{publisher}{IEEE}, \bibinfo{pages}{57--64}.
\newblock


\bibitem[Zakaria et~al\mbox{.}(2022)]%
        {zakaria2022procedural}
\bibfield{author}{\bibinfo{person}{Yahia Zakaria}, \bibinfo{person}{Magda
  Fayek}, {and} \bibinfo{person}{Mayada Hadhoud}.}
  \bibinfo{year}{2022}\natexlab{}.
\newblock \showarticletitle{Procedural Level Generation for Sokoban via Deep
  Learning: An Experimental Study}.
\newblock \bibinfo{journal}{\emph{Transactions on Games}}
  (\bibinfo{year}{2022}).
\newblock


\end{thebibliography}

\end{document}